\documentclass[conference]{IEEEtran}
\usepackage{times}

\usepackage[numbers]{natbib}
\usepackage{multicol}
\usepackage[bookmarks=true]{hyperref}

\usepackage{tabularx}
\usepackage{booktabs}
\usepackage{graphicx}
\usepackage{subcaption}
\usepackage{multirow}

\usepackage{algorithm}
\usepackage[noend]{algpseudocode}

\algnewcommand{\LineComment}[1]{\State \textcolor[rgb]{0, 0, 0.8}{$\triangleright$ #1}}

\usepackage{amsmath}
\usepackage{amsfonts}

\newcommand{\gputamp}{cuTAMP}

\usepackage{amsmath}
\usepackage{amssymb}
\usepackage[normalem]{ulem}
\usepackage{wrapfig}
\usepackage{bm}
\usepackage{xcolor}
\usepackage{soul}

\newcommand{\proc}[1]{\textsc{#1}}
\newcommand{\kw}[1]{{\bf #1}}

\newcommand{\pddl}[1]{{\texttt{#1}}} %
\newcommand{\xparam}[1]{\mathbf{x}_{\text{param}(#1)}}

\usepackage{amsthm}
\theoremstyle{plain}\newtheorem{thm}{Theorem}
\theoremstyle{definition}\newtheorem{defn}{Definition}
\theoremstyle{plain}
\theoremstyle{plain}

\usepackage{listings}
\colorlet{myyellow}{yellow!42}
\sethlcolor{myyellow}

\newif\ifcomments
\commentsfalse

\ifcomments
  \newcommand{\caelan}[1]{\textcolor{blue}{(CG: #1)}}
  \newcommand{\will}[1]{\textcolor{magenta}{(WS: #1)}}
  \newcommand{\tlp}[1]{\textcolor{orange}{(WS: TLP says: #1)}}
  \newcommand{\nishanth}[1]{\textcolor{cyan}{(NK: #1)}}
  \newcommand{\tucker}[1]{\textcolor{purple}{(TRH: #1)}}
  \newcommand{\lpk}[1]{\textcolor{orange}{LPK: #1}}
  \newcommand{\todo}[1]{\textcolor{red}{TODO: #1}}
\else
  \newcommand{\caelan}[1]{}
  \newcommand{\will}[1]{}
  \newcommand{\tlp}[1]{}
  \newcommand{\nishanth}[1]{}
  \newcommand{\tucker}[1]{}
  \newcommand{\lpk}[1]{}
  \newcommand{\todo}[1]{}
\fi

\definecolor{deepblue}{rgb}{0,0,0.5}
\definecolor{deepred}{rgb}{0.6,0,0}
\definecolor{magenta}{rgb}{1.0,0,1.0}
\definecolor{deepgreen}{rgb}{0,0.5,0}
\definecolor{textblue}{rgb}{.2,.2,.7}
\definecolor{textred}{rgb}{0.54,0,0}
\definecolor{textgreen}{rgb}{0,0.43,0}
\definecolor{es-blue}{rgb}{0.05,0.62,1}
\definecolor{bluelink}{rgb}{0,0.35,0.9}

\newcommand\blfootnote[1]{%
  \begin{NoHyper}%
  \renewcommand\thefootnote{}\footnote{#1}%
  \addtocounter{footnote}{-1}%
  \end{NoHyper}%
}

\begin{document}

\title{Differentiable GPU-Parallelized\\
Task and Motion Planning}

\author{
    \authorblockN{
        William Shen\textsuperscript{1,2*},
        Caelan Garrett\textsuperscript{2},
        Nishanth Kumar\textsuperscript{1,2},
        Ankit Goyal\textsuperscript{2},
        Tucker Hermans\textsuperscript{2,3},\\
        Leslie Pack Kaelbling\textsuperscript{1},
        Tom\'as Lozano-P\'erez\textsuperscript{1},
        Fabio Ramos\textsuperscript{2,4}
    }
    \vspace{0.5em}
    \authorblockA{
        \textsuperscript{1}MIT CSAIL,
        \textsuperscript{2}NVIDIA Research,
        \textsuperscript{3}University of Utah,
        \textsuperscript{4}University of Sydney
    }
}

\makeatletter
\let\@oldmaketitle\@maketitle
\renewcommand{\@maketitle}{\@oldmaketitle
\vspace{2.5pt}
\centering
\includegraphics[width=\linewidth]{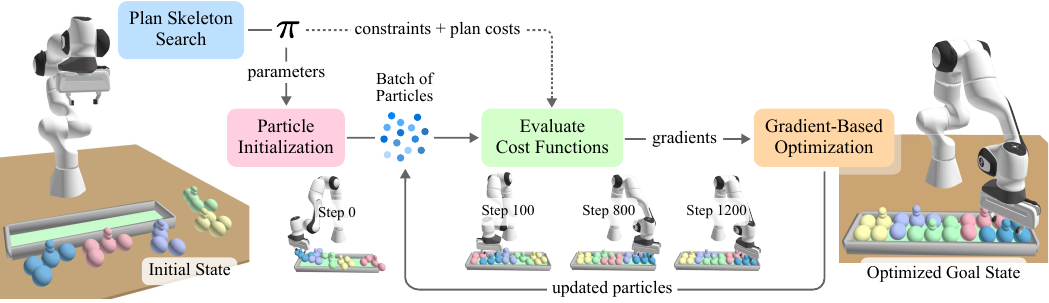}
\captionof{figure}{
\textbf{\gputamp{} Overview}. 
\gputamp{} frames TAMP as a backtracking bilevel search over plan skeletons (Sec.~\ref{sec:overview}). Each skeleton \(\pi\) induces a continuous Constraint Satisfaction Problem that defines the structure of a particle (parameters) and cost functions (constraints and plan costs).
These particles are optimized in parallel by evaluating their costs with differentiable cost functions (Eq.~\ref{eq:batched-particle-cost}), allowing gradient-based optimizers to iteratively update them towards satisfying solutions (Sec.~\ref{sec:diff-opt}).
}
\label{fig:teaser}
\vspace{-5px}
}

\makeatother

\maketitle

\begin{abstract}
Planning long-horizon robot manipulation requires making discrete decisions about which objects to interact with and continuous decisions about how to interact with them.
A robot planner must select grasps, placements, and motions that are feasible and safe.
This class of problems falls under Task and Motion Planning (TAMP) and poses significant computational challenges in terms of algorithm runtime and solution quality, particularly when the solution space is highly constrained.
To address these challenges, we propose a new bilevel TAMP algorithm that leverages GPU parallelism to efficiently explore thousands of candidate continuous solutions simultaneously.
Our approach uses GPU parallelism to sample an initial batch of solution seeds for a plan skeleton and to apply differentiable optimization on this batch to satisfy plan constraints and minimize solution cost with respect to soft objectives.
We demonstrate that our algorithm can effectively solve highly constrained problems with non-convex constraints in just seconds, substantially outperforming serial TAMP approaches, and validate our approach on multiple real-world robots.
Project website and code: {\color{bluelink}\href{https://cutamp.github.io}{cutamp.github.io}}
\blfootnote{
    \textsuperscript{*}Work partially conducted during internship at NVIDIA.
    Correspondence to: William Shen (willshen@mit.edu), Caelan Garrett (cgarrett@nvidia.com).
}
\end{abstract}

\IEEEpeerreviewmaketitle

\begin{figure*}
\centering
\begin{minipage}[t]{0.475\textwidth}
    \centering
    \includegraphics[width=\linewidth]{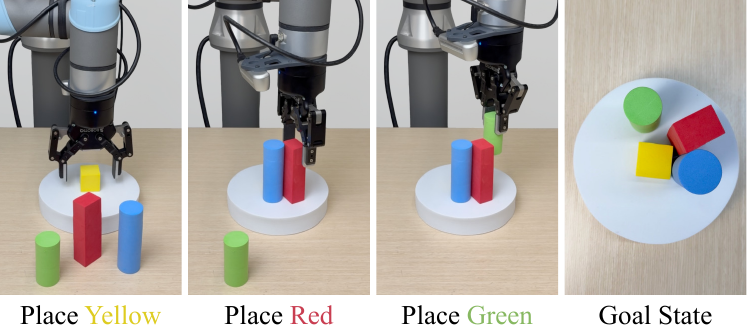}
    \caption{\textbf{Object Packing with a UR5.} The objective is to place all objects onto the white region while minimizing the distance between them. 
    The final state achieves a tight packing with successful reduction of the goal cost.
    }
    \label{fig:ur5-min-obj-dist}
\end{minipage}
\hfill
\begin{minipage}[t]{0.50\textwidth}
    \centering
    \includegraphics[width=\linewidth]{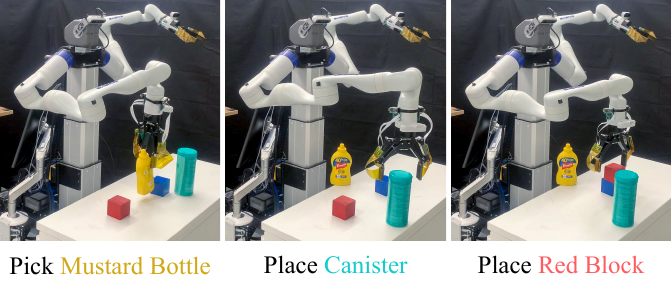}
    \caption{
        \textbf{Block Stacking with a Kinova Arm.} The objective is to stack the red block on the blue block. However, the mustard bottle and canister obstruct all placements. \gputamp{} reasons to move these objects out of the way before placing the red block.
    }
    \label{fig:rummy-obstruction}
\end{minipage}
\vspace{-10px}
\end{figure*}

\section{Introduction}
Task and Motion Planning (TAMP) enables robots to plan long-horizon manipulation through integrated reasoning about sequences of discrete action types, such as pick, place, or press, and continuous action parameter values, such as grasps, placements, and trajectories~\cite{garrett2021survey}.
TAMP planners have demonstrated remarkable generality in complex tasks including object rearrangement~\cite{garrett2020PDDLStream}, multi-arm assembly~\cite{chen2022cooperative}, and cooking a meal~\cite{yang2022sequence}.
However, TAMP problems become increasingly challenging to solve efficiently as the horizon and action space increase, and the size of the set of solutions decreases due to tightly interacting constraints, e.g., kinematics and collisions. %

A popular family of TAMP algorithms solve problems by first searching over discrete action sequences, also known as {\em plan skeletons}, and then searching for continuous action parameter values that satisfy the collective action {\em constraints} that govern legal parameter values.
Each candidate plan skeleton 
induces a continuous {\em Constraint Satisfaction Problem} (CSP), which TAMP algorithms typically solve using a mixture of compositional {\em sampling} and joint {\em optimization} techniques, with each having their own trade-offs~\cite{garrett2021survey}.

Sampling-based approaches to TAMP disconnect the parameters by generating samples for each independently using hand-engineered~\cite{kaelbling2011hpn,srivastava2014combined,garrett2018sampling}, projection-based~\cite{vega2020asymptotically}, or learned generators~\cite{fang2023dimsam,yang2023diffusion}, and then combining them through composition and rejection.
Because the parameters only interact through rejection sampling when evaluating constraints, many samples are often needed to satisfy problems where the constraints interact, such as tight packing problems (Figure~\ref{fig:teaser}).
Optimization-based TAMP approaches, on the other hand, represent constraints as analytic functions in a mathematical program
and solve for the continuous parameters by applying first- or second-order gradient descent~\cite{garrett2021survey,zhao2024survey}.
However, these constrained mathematical programs are highly non-convex with many local optima, making it challenging to find even a feasible solution from random parameter initializations.

We present \gputamp{}, the first GPU-parallelized TAMP planner.
\gputamp{} enables massively parallel exploration of TAMP solutions by combining ideas from sampling-based and optimization-based TAMP with GPU acceleration, going beyond prior serial algorithms.
We treat TAMP constraint satisfaction as simultaneous differentiable optimization over a batch of \textit{particles}, representing thousands of candidate solutions.
This allows us to maintain the interdependence between continuous parameters by jointly optimizing them.
To initialize the particles, we leverage parallelized samplers that solve \textit{constraint subgraphs}, composing their generations to populate particles near the solution manifold while ensuring good coverage of parameter space.
We demonstrate that when massively parallelized, \gputamp{} can effectively solve highly constrained TAMP problems.
Our approach inherits the locality of gradient descent and explores multiple basins through compositional sampling, increasing the likelihood of finding the global optima.
Although we focus on GPU acceleration, our method applies to other forms of parallel computation.

We evaluate \gputamp{} on a diverse range of TAMP problems of varying difficulty and highlight the benefits of GPU parallelism.
By scaling the number of particles, we achieve significant improvements in the number of satisfying solutions, algorithm runtime, and solution quality.
For highly constrained problems that baselines fail to solve, \gputamp{} finds solutions in just seconds.
We deploy our algorithm on a real UR5 and Kinova arm and showcase its fast planning capabilities for long-horizon manipulation problems (Figures~\ref{fig:ur5-min-obj-dist} and~\ref{fig:rummy-obstruction}).
Code and videos are available on our website: {\color{bluelink}\href{https://cutamp.github.io}{cutamp.github.io}.}

\section{Related Work}

\textbf{Parallelized Motion Planning.} Early algorithms for parallelized motion planning used multiprocessing~\cite{henrich1997fast} during primarily embarrassingly parallelizable operations, for example, when computing an explicit representation of the robot's configuration space~\cite{lozanoperez1991parallel}.
More recent algorithms leverage vectorization~\cite{thomason2024motions} or GPU-acceleration~\cite{kider2010high,pan2012gpu} to implement primitive motion planning operations, such as forward kinematics and collision checking.
Our work is most closely related to cuRobo~\cite{sundaralingam2023curobo}, which leverages GPU-acceleration in two phases: first, during a sampling-based Probabilistic Roadmap (PRM)~\cite{kavraki1996probabilistic} phase that generates candidate paths, and second, during a trajectory optimization phase seeded from these paths that minimizes trajectory duration subject to dynamical limits.
In contrast, \gputamp{} addresses the broader problem of TAMP (i.e., manipulation planning), which requires reasoning about grasps, contacts, and placements in addition to motions.

\textbf{Sampling-Based TAMP.}  Sampling-based TAMP algorithms handle the continuous decision-making within TAMP through discretization and composition.
They generate values that satisfy specific constraints, such as grasp and placement stability constraints, and intersect them with additional constraints, such as kinematic and collision constraints, through rejection and conditional sampling~\cite{garrett2021survey}.
Prior approaches do this by sampling a fixed problem discretization~\cite{garrett2018ffrob,dantam2018incremental}, combining generators using a custom interface layer~\cite{srivastava2014combined}, searching through the multi-modal continuous space~\cite{hauser2011randomized,vega2020asymptotically}, specifying geometric suggesters~\cite{kaelbling2011hpn}, and composing samplers using a stream specification~\cite{garrett2018sampling,garrett2020PDDLStream,khodeir2023learning,ren2024extended}.
Additionally, it is possible to cast satisfaction as an inference problem and leverage techniques like Markov chain Monte Carlo (MCMC)~\cite{toussaint2024nlp} and Stein Variational Inference~\cite{lee2023stamp}.
We leverage sampling to populate candidate \emph{particles} that are near the solution manifold, but not necessarily feasible, for gradient-based optimization.

\textbf{Optimization-Based TAMP.} In contrast to sampling-based algorithms, which leverage constraint compositionality, optimization-based TAMP algorithms directly solve for continuous parameters that jointly satisfy all plan constraints.
Although sampling-based methods can generally optimize plan cost in an anytime mode using rejection sampling, optimization-based TAMP leverages mathematical programming, often via first- and higher-order optimization.
Several approaches use off-the-shelf Mixed Integer Programming (MIP) solvers for constraint satisfaction; however, these approaches are limited to simplified TAMP problems where the continuous dynamics are linear~\cite{garrett2021sampling,quintero2023optimal} or convex~\cite{shoukry2017smc,shoukry2018smc,Fernandez-Gonzalez2018,adu2022optimal}.
Other approaches directly address the non-convex constrained optimization problem using Sequential Quadratic Programming (SQP)~\cite{hadfield2016sequential}, Augmented Lagrangian methods~\cite{toussaint2014newton,toussaint2015logic,Toussaint-RSS-18,garrett2021sampling,ortiz20222conflict,ortiz2022diverse}, and Alternating Direction Method of Multipliers (ADMM)~\cite{zhao2021sydebo}.
These methods are computationally expensive per attempt and are not guaranteed to converge to a feasible solution.
By optimizing thousands of candidates solutions in parallel, our approach is more likely to produce at least one feasible solution.

\begin{figure}[t]
\begin{footnotesize}
\begin{lstlisting}
MoveFree|$(q_1, q_2: \text{conf},\; \tau: \text{traj})$|
 |\kw{con}:| [Motion|$(q_1, \tau, q_2)$|, CFreeTraj|$(\tau)$|]
 |\kw{pre}:| [AtConf|$(q_1)$|, HandEmpty|$()$|]
 |\kw{eff}:| [AtConf|$(q_2)$|, |$\neg$|AtConf|$(q_1)$|]
 |\kw{cost}:| [TrajLength|$(\tau)$|]
\end{lstlisting}
\vspace{-0.275cm}
\begin{lstlisting}
Pick|$(o: \text{obj},\; g: \text{grasp},\; p: \text{placement},\; q: \text{conf})$|
 |\kw{con}:| [Kin|$(q, o, g, p)$|, Grasp|$(o, g)$|, CFreeHold|$(o, g, q)$|]
 |\kw{pre}:| [HandEmpty|$()$|, AtConf|$(q)$|, AtPlacement|($o, p$)|]
 |\kw{eff}:| [Holding|$(o, g)$|, |$\neg$|HandEmpty|$()$|, |$\neg$|AtPlacement|$(o, p)$|]
\end{lstlisting}
\vspace{-0.275cm}
\begin{lstlisting}
MoveHold|$(o: \text{obj},\; g: \text{grasp},\; q_1, q_2: \text{conf},\; \tau: \text{traj})$|
 |\kw{con}:| [Motion|$(q_1, \tau, q_2)$|, CFreeTrajHold|$(o, g, \tau)$|]
 |\kw{pre}:| [AtConf|$(q_1)$|, Holding|$(o, g)$|]
 |\kw{eff}:| [AtConf|$(q_2)$|, |$\neg$|AtConf|$(q_1)$|]
 |\kw{cost}:| [TrajLength|$(\tau)$|]
\end{lstlisting}
\vspace{-0.275cm}
\begin{lstlisting}
Place|$(o: \text{obj},\; g: \text{grasp},\; p: \text{placement},\; s: \text{surface},\; q: \text{conf})$|
 |\kw{con}:| [Kin|$(q, o, g, p)$|, StablePlace|$(o, p, s)$|, CFreePlace|$(o, p)$|]
 |\kw{pre}:| [Holding|$(o, g)$|, AtConf|$(q)$|]
 |\kw{eff}:| [HandEmpty|$()$|,AtPlacement|$(o, p)$|,On|$(o, s)$|,|$\neg$|Holding|$(o, g)$|]
\end{lstlisting}
\end{footnotesize}
\vspace{-0.15cm}
\captionof{lstlisting}{
\textbf{Parametrized actions for pick and place tasks.} We list the most important constraints for simplicity of presentation. CFree is an abbreviation for \underline{C}ollision \underline{Free}. 
}
\label{lst:actions}
\vspace{-10px}
\end{figure}

\section{Problem Formulation}
\label{sec:problem-formulation}
Our approach is generally applicable to long-horizon decision-making problems with both discrete and continuous parameters, such as assembly line design, smart power grid management, and programming video game non-playable characters. We focus on solving TAMP problems. 

Let a TAMP problem be a tuple \(\Pi = \langle \mathcal{A}, s_0,  S_*\rangle\), where \(\mathcal{A}\) is a set of \emph{parametrized actions}, \(s_0\) is the initial state, and \(S_*\) is the set of goal states.
We represent states and actions using a PDDL-style (Planning Domain Definition Language)~\cite{mcdermott1998pddl} action language, where states are comprised of Boolean variables corresponding to logical propositions. %
For example, the \pddl{AtPlacement}($o, p$) {\em predicate} holds if object $o$ is currently at placement pose $p$.
Each parametrized action \(a \in \mathcal{A}\) accepts parameters \(\mathbf{x}_a = (x_1, \dots, x_n)\), which may include both discrete and continuous values, and consists of: %
\begin{itemize}
    \item Constraints (\textbf{con}) on its parameters, which must all be satisfied in order for the action to be valid in some state. We assume that the constraints are equality or inequality constraints on a differentiable real-valued function, denoted as \(J_c\) for each constraint \(c \in \textbf{con}(a)\) in action \(a\).
    \item Preconditions (\textbf{pre}), which must all be true for the action to be executed in a given state.
    \item Effects (\textbf{eff}), which describe propositions that become true or false after executing the action.
    \item Costs (\textbf{cost}) on the parameters, which we aim to reduce.
\end{itemize}

For example, consider the \texttt{Pick} action in Listing~\ref{lst:actions} for grasping object \(o\), with the grasp \(g\), when at pose \(p\), and corresponding robot configuration \(q\).
Its preconditions are 1) $\pddl{HandEmpty}()$ -- the robot's hand must be empty, 2) $\pddl{AtConf}(q)$ -- the robot must be at configuration \(q\), and 3) $\pddl{AtPlacement}(o, p)$ -- object \(o\) must be at placement pose \(p\).
As a result of executing the \texttt{Pick} action, $\pddl{AtGrasp}(o, g)$ -- the robot holds object \(o\) with grasp \(g\) -- is now true, but $\pddl{HandEmpty}()$ and $\pddl{AtPlacement}(o, p)$ are now false.
To execute the action, we require that the kinematic constraint \(\pddl{Kin}(q, o, g, p)\) is satisfied (i.e., \(\proc{FK}(q) = p \cdot g\) where FK is forward kinematics), \(g\) is a valid grasp for object \(o\) as required by \(\pddl{Grasp}(o, g)\), and the grasp is \underline{C}ollision-\underline{Free} (CFree) at configuration \(q\) required by \(\pddl{CFreeHold}(o, g, q)\).
We provide additional \pddl{MoveFree}, \pddl{MoveHold}, and \pddl{Place} actions in Listing~\ref{lst:actions}.
The complete description of the constraints and actions considered may be found in Appendix~\ref{app:problem-formulation}.

The objective of the TAMP system is to find a plan skeleton \(\pi = (a_1, \dots, a_n)\), a sequence of actions,
with valid parameter assignments \(\{\mathbf{x}_a\mid a \in \pi\}\) such that when applied from the initial state \(s_0\), it produces a goal state \(s \in S_*\) while minimizing the overall cost. 

\textbf{Running Example.} Consider the following skeleton for placing object $\pddl{red}$ on surface $\pddl{table}$, where constant continuous parameters are bolded:

\vspace{-1.1em}
\begin{small}%
\begin{align*}
\pi = [&\pddl{MoveFree}(\bm{q_0}, q_1, \tau_1),\; \pddl{Pick}(\pddl{red}, g, \bm{p_0}, q_1), \\
&\pddl{MoveHold}(\pddl{red}, g, q_1, q_2, \tau_2),\; \pddl{Place}(\pddl{red}, g, p_1, \pddl{table}, q_2)].
\end{align*}
\end{small}

Parameters in a plan skeleton may be shared across actions. 
In order to \texttt{Pick} object \pddl{red} at configuration \(q_1\), the robot must first use \texttt{MoveFree} to move from its initial configuration \(\bm{q_0}\) to \(q_1\). 
A key challenge in TAMP is that continuous constraints may restrict the set of viable plan skeletons.
For example, a \pddl{blue} object may initially obstruct the \pddl{red} object, causing \pddl{CFreeHold} in \pddl{Pick} to be false.
This skeleton would admit no solutions as long as \pddl{blue} is at its initial placement.

\begin{figure}[t]
\centering
\begin{subfigure}{0.493\linewidth}
    \centering
    \includegraphics[width=\textwidth]{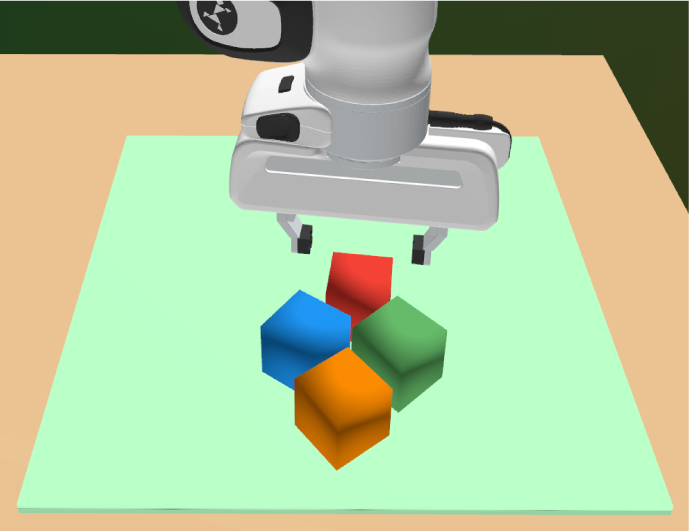}
    \caption{\gputamp{} with goal costs}
    \label{fig:obstacle-blocks-soft-cutamp}
\end{subfigure}
\hfill
\begin{subfigure}{0.493\linewidth}
    \centering
    \includegraphics[width=\textwidth]{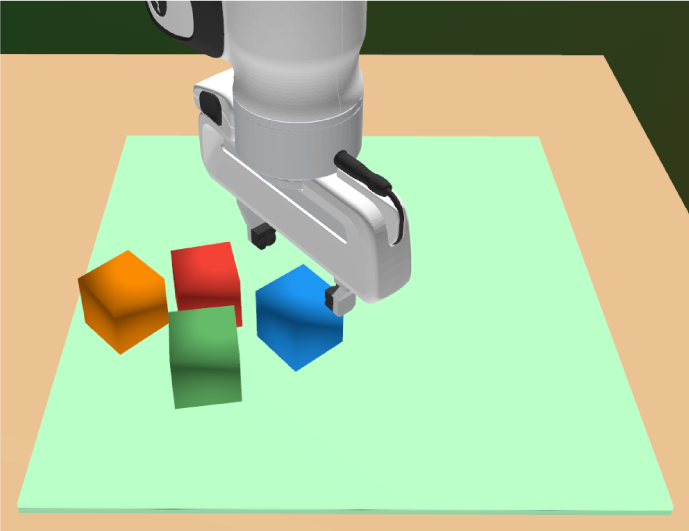}
    \caption{Parallelized Sampling}
\end{subfigure}
\caption{
    \textbf{Minimizing Distance between Objects.} The state after executing the best particle. (a) \gputamp{} achieves significantly lower cost compared to (b) parallelized sampling.
}
\label{fig:obstacle-blocks-soft}
\vspace{-10px}
\end{figure}

\textbf{Goal Costs.}
In some tasks, the objective is to reach a state that additionally minimizes a cost function, such as the goal distance between objects (Figure~\ref{fig:obstacle-blocks-soft}).
We support this by treating costs on the goal state as a dummy action with costs that are appended as the final action in candidate plan skeletons.
In our example, this action for three objects is:

\vspace{-0.05cm}%
\begin{small}
\begin{lstlisting}[breaklines=true]
MinimizeObjDist|$(o_1, o_2, o_3: \text{obj},\; p_1, p_2, p_3: \text{placement})$|
 |\kw{pre}:| [AtPlacement|$(o_1, p_1)$,\; AtPlacement|$(o_2, p_2)$|,\; AtPlacement|$(o_3, p_3)$]
 |\kw{cost}:| [Dist|$(p_1, p_2)$|, Dist|$(p_2, p_3)$|, Dist|$(p_1, p_3)$|]
\end{lstlisting}
\end{small}

\section{\gputamp{} Overview}
\label{sec:overview}

\gputamp{} is a \emph{sequence-then-satisfy} approach to TAMP~\cite{garrett2021survey}: it first searches over plan skeletons and then searches over continuous action parameter values for each of the actions within that skeleton.
Specifically, \gputamp{} first generates a candidate skeleton and then uses massively parallelized differentiable optimization to solve the induced constraint satisfaction problem (CSP) (Section~\ref{sec:csp}).
Since any particular skeleton might induce a CSP with no feasible solution, \gputamp{} \emph{backtracks} to attempt different skeletons until it finds a feasible one.
To ensure efficient optimization,
\gputamp{} invokes sampling methods to initialize a batch of \textit{particles}, representing candidate \textit{continuous} solutions to the CSP (Section~\ref{sec:particle-initialization}).
Finally, to minimize backtracking, \gputamp{} derives a plan feasibility heuristic to guide the discrete search over skeletons (Section~\ref{sec:search-over-skeletons}).
In Section~\ref{appendix:cutamp-prob-comp}, we prove that \gputamp{} is {\em probabilistically complete}.

The \gputamp{} algorithm is listed in Algorithm~\ref{alg:gpu-tamp}.
In Stage 1, it searches for \(N_s\) initial plan skeletons using task planning, which we perform using a forward 
best-first search that introduces new parameters for continuous variables whenever it expands a node in the search tree. 
For each skeleton \(\pi_i\), it performs particle initialization to get a batch of particles \(\mathcal{P}_i\) (Sec.~\ref{sec:particle-initialization}), and computes the \textsc{PlanHeuristic} (Sec.~\ref{sec:search-over-skeletons}).
These candidate skeletons and particles are appended to a priority queue \(Q\), which is ordered by the heuristic values.
In Stage 2, it iteratively selects plan skeletons and their particles from \(Q\) and optimizes the particles using differentiable optimization (Sec.~\ref{sec:diff-opt}) until a stopping criterion is met (e.g., time limit or number of steps).
After optimizing the particles, it evaluates whether the overall stopping conditions have been met, such as finding at least one satisfying solution or reaching a user-defined cost value.
If the conditions are met, it returns all satisfying solutions.
Otherwise, it recomputes the plan heuristic on the optimized particles and samples additional skeletons and particles to add to the priority queue.

In the sections that follow, we discuss each component of \gputamp{} in detail. 
At the core of our approach is our use of parallelized differentiable optimization for solving CSPs.
We thus present this first in Section~\ref{sec:csp}.
Like most optimization-based approaches, ours is sensitive to initialization. \gputamp{} addresses this challenge with particle initialization strategies, which we present next in Section~\ref{sec:particle-initialization}.
Finally, we describe our approach for guiding the discrete search over plan skeletons, which leverages the particle initializations, in Section~\ref{sec:search-over-skeletons}.

\begin{algorithm}[t]
\begin{footnotesize}
\caption{\gputamp{} Algorithm}
\label{alg:gpu-tamp}
\begin{algorithmic}[1]
\Require TAMP problem \(\Pi\), \(N_b\) particle batch size, \(N_s\) number of new plan skeletons
\State \(Q \gets [ \ ]\)  \Comment{Initialize priority queue}
\For{\(i = 1, \dots, N_s\)} \Comment{Stage 1: Initialize plans and particles.}
    \State \(\pi_i \gets \textsc{SearchPlanSkeleton}(\Pi) \)
    \State \(\mathcal{P}_i \gets \textsc{InitializeParticles}(\pi_i, N_b)\) \Comment{Parallelized Sampling}
    \State \(h_i \gets \textsc{PlanHeuristic}(\pi_i, \mathcal{P}_i)\)
    \State \(\textsc{Push}(Q, (h_i, \pi_i, \mathcal{P}_i))\)
\EndFor
\While{\(Q \neq [ \ ]\)} \Comment{Stage 2: Loop over skeletons and optimize.}
    \State \((h, \pi, \mathcal{P}) \gets \textsc{Pop}(Q)\)    
    \State \(\mathcal{P}' \gets \textsc{OptimizeParticles}(\pi, \mathcal{P})\) \Comment{Parallelized Optimization}
    \If{\(\textsc{IsGoalSatisfied}(\pi, \mathcal{P}')\)}
        \State \Return \(\pi, \textsc{GetSatisfyingParticles}(\pi, \mathcal{P}')\)
    \EndIf
    \For{\(i = 1, \dots, N_s\)} \Comment{Sample additional plan skeletons.}
    \State \(\pi_{\text{new}} \gets \textsc{SearchPlanSkeleton}(\Pi)\)
        \State \(\mathcal{P}_{\text{new}} \gets \textsc{InitializeParticles}(\pi_{\text{new}}, N_b)\)
        \State \(h_{\text{new}} \gets \textsc{PlanHeuristic}(\pi_{\text{new}}, \mathcal{P}_{\text{new}})\)
        \State \(\textsc{Push}(Q, (h_{\text{new}}, \pi_{\text{new}}, \mathcal{P}_{\text{new}}))\)
    \EndFor
    \State \(h' \gets \textsc{PlanHeuristic}(\pi, \mathcal{P}')\)  %
    \State \(\textsc{Push}(Q, (h', \pi, \mathcal{P}'))\)  \Comment{Add back to queue}
\EndWhile
\State \Return \(\textsc{Failure}\)
\end{algorithmic}
\end{footnotesize}
\end{algorithm}

\section{Parallelized Constraint Satisfaction}
\label{sec:csp}

\begin{figure*}[t]
    \centering
    \begin{minipage}[t]{0.375\textwidth}
        \centering
        \includegraphics[width=\linewidth]{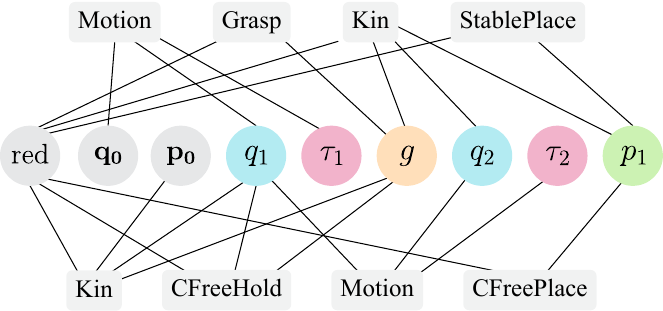}
        \caption{\textbf{Example Constraint Network.} Variables (round nodes) are connected to each other via constraints (rectangular nodes). We omit \pddl{CFreeTraj} constraints for simplicity.}
        \label{fig:constraint-network}
    \end{minipage}
    \hfill
    \begin{minipage}[t]{0.6\textwidth}
        \centering
        \includegraphics[width=\linewidth]{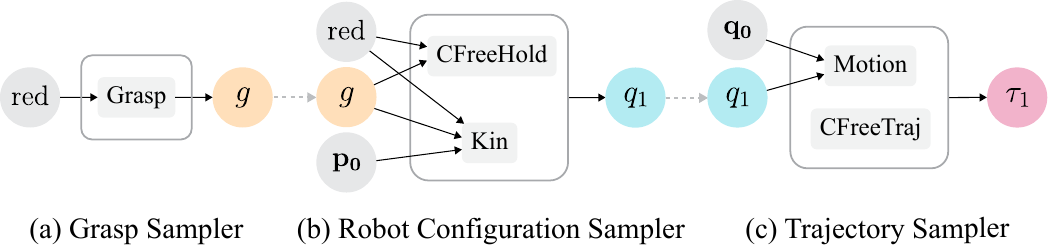}
        \caption{\textbf{Example Sampling Networks.} Samplers solve subgraphs of the constraint network (Fig.~\ref{fig:constraint-network}). Solid arrows indicate the input and output parameters for each sampler. Dotted arrows represent dependencies between the samplers, defining the order in which they can be \emph{composed}.
        }
        \label{fig:individual-sampling-networks}
    \end{minipage}
    \vspace{-10px}
\end{figure*}

A candidate plan skeleton \(\pi = (a_1, \dots, a_n)\) induces a continuous Constraint Satisfaction Problem (CSP), where the goal is to assign values to the continuous parameters such that all the constraints across \(a \in \pi\), are satisfied.
We denote the set of constraints across \(\pi\) as \(\textbf{con}(\pi) = \bigcup_{a \in \pi} \textbf{con}(a)\), and the cost functions as \(\textbf{cost}(\pi) = \bigcup_{a \in \pi} \textbf{cost}(a)\).

A CSP can be visualized as a constraint network, where nodes represent variables and constraints, and edges connect variables via the constraints.
Figure~\ref{fig:constraint-network} depicts the constraint network for the running example (Sec.~\ref{sec:problem-formulation}) for placing object \pddl{red} on surface \pddl{table}, where the free variables are \(q_1\), \(\tau_1\), \(g\), \(q_2\), \(\tau_2\), and \(p_1\), while the set of constraints \(\textbf{con}(\pi)\) are:%

\vspace{-1.5em}
\begin{small}
\begin{align*}
\{\;
&\pddl{Motion}(\bm{q_0}, \tau_1, q_1),\;\; \pddl{CFreeTraj}(\tau_1),\;\; \pddl{Kin}(q_1, \pddl{red}, g, \bm{p_0}),\\
&\pddl{Grasp}(\pddl{red}, g),\;\; \pddl{CFreeHold}(\pddl{red}, g, q_1),\;\; \pddl{Motion}(q_1, \tau_2, q_2),\\
&\pddl{CFreeTrajHold}(\pddl{red}, g, \tau_2),\;\; \pddl{Kin}(q_2, \pddl{red}, g, p_1),\\
&\pddl{StablePlace}(\pddl{red}, p_1, \pddl{table})\},\;\;\pddl{CFreePlace}(\pddl{red}, p_1)\;\}.
\end{align*}
\end{small}

\subsection{Constraint Satisfaction via Optimization}
\label{sec:csp-as-optimization}

We solve the CSP induced by skeleton \(\pi\) by reducing it to an unconstrained optimization problem involving the real-valued functions comprising each constraint and the cost functions across \(\pi\).
Let a parameter {\em particle} \(\mathbf{x} = (x_1, x_2, \dots, x_{N_v})\) be an assignment to the \(N_v\) continuous variables in \(\pi\).
For each constraint \(c \in \textbf{con}(\pi)\), we denote its differentiable real-valued function as \(J_{c}\) and tolerance as \(\epsilon_{c}\).
A constraint \(c\) is satisfied if \(J_{c}(\xparam{c}) \leq \epsilon_{c}\), where \(\xparam{c}\) denotes the subset of parameters in \(\mathbf{x}\) that are relevant to \(c\).
In our running example, the kinematics constraint \pddl{Kin}($q_1$, \pddl{red}, $g$, $\bm{p_0}$) performs forward kinematics on the robot configuration $q_1$ and returns the pose error relative to the target pose $\bm{p_0}\cdot g$ for object \pddl{red}.
A particle \(\mathbf{x}\) is \textit{satisfying} if it satisfies all the constraints in \(\pi\), and hence forms a valid solution to the CSP.
Finding such a satisfying particle corresponds to solving the following mathematical program:%
\begin{equation}
\label{eq:csp-program}
\begin{aligned}
\min_{\mathbf{x}} \quad & \sum_{c' \in \textbf{cost}(\pi)} c'(\xparam{c'}) \\
\text{subject to} \quad & J_c(\xparam{c}) \leq \epsilon_c \qquad \forall c \in \textbf{con}(\pi).
\end{aligned}
\end{equation}
Each plan skeleton induces a mathematical program, where the variables and costs are determined by the parameters, constraints, and costs of its actions.
We solve the program by relaxing the hard constraints into soft costs, enabling the use of unconstrained optimization.
The objective function is a weighted sum of the costs from the hard constraints and the plan costs, which may be viewed as soft constraints:
\begin{equation}
\begin{aligned}
\label{eq:particle-cost}
\min_\mathbf{x} \ \mathcal{J}(\mathbf{x}) = 
    &\underbrace{
        \sum_{c \in \textbf{con}(\pi)} 
        \lambda_{c} \cdot J_{c}(\xparam{c}) 
    }_{\text{hard constraint costs}} \\
    &+
    \underbrace{
        \sum_{c' \in \textbf{cost}(\pi)} 
        \lambda_{c'} \cdot c'(\xparam{c'})
    }_{\text{soft action costs}},
\end{aligned}%
\end{equation}%
where \(\lambda_c\) is the weight (constant penalty) for the corresponding constraint or cost \(c\), which allows us to balance their influence during the optimization process.
We can check whether a particle \(\mathbf{x}\) satisfies the CSP by evaluating:%
\begin{equation}
\label{eq:satisfying-particle}
\bigwedge_{c \in \textbf{con}(\pi)} J_c(\mathbf{x}_{\text{param}(c)}) \leq \epsilon_c.
\end{equation}

\subsection{Parallelized Differentiable Optimization}
\label{sec:diff-opt}

A key contribution of our work is to exploit parallelism to apply differentiable optimization and explore thousands of parameter particles simultaneously.
We first denote a batch of \(N_b\) parameter particles as \(\mathcal{P} =(\mathbf{x}_1, \dots, \mathbf{x}_{N_b})\).
The objective in our unconstrained optimization problem is now to minimize the mean cost over all particles in \(\mathcal{P}\):%
\begin{equation}
\label{eq:batched-particle-cost}
    \min_{\mathcal{P}} \ \mathcal{J}_{\text{batch}}(\mathcal{P}) = \frac{1}{N_b} \sum_{\mathbf{x} \in \mathcal{P}} \mathcal{J}(\mathbf{x}).
\end{equation}%

\textbf{Gradient-Based Optimization.}
We solve the unconstrained optimization problem in Equation~\ref{eq:batched-particle-cost} by iteratively updating the particles \(\mathcal{P}\) using a gradient-based optimizer.
At each optimization step, we compute the cost function across the batch of particles.
Since the cost function is differentiable, we can compute the gradients of the cost with respect to the particles.
These gradients are then used by Adam~\cite{kingma2014adam}, a stochastic first-order optimizer, to update the parameters within each particle.
Our results in Section~\ref{sec:experiments} demonstrate that this approach performs remarkably well \textit{even} with simple unconstrained optimization using Adam;
however, our approach is compatible with more complex optimizers including augmented Lagrangians~\cite{platt1987constrained,landry2019differentiable,heiden2022probabilistic}, coordinate descent~\cite{wright2015coordinate}, and second-order optimizers~\cite{liu1989limited}. %
We repeat gradient-descent updates until a stopping criterion has been met, such as a maximum optimization time or finding a satisfying particle within the batch.
We check whether a particle is satisfying by comparing whether the costs corresponding to the plan skeleton's constraints falls within the defined tolerances (Eq.~\ref{eq:satisfying-particle}).

\textbf{Parallelizing on GPUs.} 
To efficiently optimize a batch of particles, we avoid the inefficient summation in Eq.~\ref{eq:batched-particle-cost} by first stacking the assignments of each continuous variable \(x_i\) across the batch of \(N_b\) particles into a matrix \(\mathbf{X}_i\).
For example, if \(x_i \in \mathbb{R}^7\) represents a 7-DOF robot configuration, then \(\mathbf{X}_i \in \mathbb{R}^{N_b \times 7}\).
In order to compute the cost in Eq.~\ref{eq:particle-cost} across batches of particles simultaneously, we implement vectorized versions of the cost functions using PyTorch~\cite{paszke2019pytorch}.
These cost functions are differentiable via automatic differentiation, allowing us to compute gradients in parallel.
We also leverage differentiable collision checkers and kinematics models within cuRobo~\cite{sundaralingam2023curobo}, which include custom CUDA kernels for both the forward pass and gradient computations.
We use the kinematics model to compute pose errors for kinematic constraints.
Collision checking approximates the geometry of the robot and movable objects as spheres, enabling massively GPU-accelerated collision checking against static obstacles represented as oriented bounding boxes, meshes, or signed distance fields.
We use these collision checkers, which provide informative and smooth gradients, in our cost functions for checking the collision-free and self-collision constraints.
See Appendix Listing~\ref{appendix:example-cost-functions} for example cost functions in Python.

\section{Particle Initialization}
\label{sec:particle-initialization}
A key desideratum for solving the highly non-convex optimization problem presented above in Section~\ref{sec:diff-opt} is avoiding local minima.
Towards this, \gputamp{} implements a novel strategy based on compositional sampling to initialize the particles \(\mathcal{P}\) before optimization is run.
One common strategy in optimization-based TAMP is to restart with random initializations when stuck in local minima~\cite{toussaint2024nlp}.
However, as we show in Section~\ref{sec:experiments}, targeted initialization via conditional sampling skips the early stages of optimization that must move near the solution manifold.
This allows us to jump straight to improving the particles according to the constraints jointly.

Let a \textit{parallelized sampler} be a function that takes one or more constraints, possibly involving different constraint types, as input and generates a batch of \(N_b\) assignments to the free parameters involved in those constraints.
A \textit{conditional parallelized sampler} accepts assignments for some of the free parameters as additional input.
A sampler's objective is to find assignments that satisfy its constraints; however, this is only possible when the constraints admit satisfying assignments.

Our initialization strategy is to \emph{compose} the generations of multiple samplers, where each sampler solves a subgraph of the constraint network.
This provides an initialization for the entire batch of particles \(\mathcal{P}\), corresponding to \(N_b\) candidate solutions to the CSP.
This composition forms a sampling network (also known as a computation graph), an orientation of the edges in a constraint network that transforms it into a directed acyclic graph (DAG)~\cite{garrett2018sampling}.

\textbf{Concrete Example.} Consider again the plan skeleton from our running example in Section~\ref{sec:problem-formulation}, for which the constraint network is depicted in Figure~\ref{fig:constraint-network}. 
We outline our use of samplers to initialize particles for the first two actions in the skeleton: 1) \(\pddl{MoveFree}(\bm{q_0}, q_1, \tau_1)\) and 2) \(\pddl{Pick}(\pddl{red}, g, \bm{p_0}, q_1)\), where the free variables are \(q_1\), \(\tau_1\), \(g\) and \(q_1\).
Similar to the approach in~\cite{garrett2018sampling}, we provide the following samplers:

\begin{enumerate}
    \item A 6-DOF grasp sampler that takes the \pddl{Grasp}(\pddl{red}, $g$) constraint as input and generates grasps \(\mathbf{G}\) in the object frame (Figure~\ref{fig:individual-sampling-networks}a).
    \item A conditional sampler for robot configurations that takes the grasp samples \(\mathbf{G}\) and constraints \pddl{Kin}($q_1$, \pddl{red}, $g$, $\bm{p_0}$) and \pddl{CFreeHold}(\pddl{red}, $g$, $q_1$) as input (Figure~\ref{fig:individual-sampling-networks}b). The sampler uses the parallelized inverse kinematics solver from cuRobo~\cite{sundaralingam2023curobo} to solve for 7-DOF joint positions \(\mathbf{Q}_1 \in \mathbb{R}^{N_b \times 7}\) conditioned on the target end-effector poses derived from the grasps \(\mathbf{G}\).
    \item A conditional trajectory sampler for \(\tau_1\) that takes the configurations \(\mathbf{Q}_1\) and the \pddl{Motion}($\bm{q_0}$, $\tau_1$, $q_1$) and \pddl{CFreeTraj}($\tau_1$) constraints as input (Figure~\ref{fig:individual-sampling-networks}c). The sampler could then linearly interpolate between the initial configuration \(\bm{q_0}\) and \(\mathbf{Q}_1\) or sample from a learned diffusion model~\cite{huang2024diffusionseeder}.
\end{enumerate}
\noindent
By composing these samples, we obtain an initialization for the entire batch of particles \(\mathcal{P}\) which corresponds to \(N_b\) candidate solutions to the CSP.

\textbf{Reusing Samples across Skeletons.}
Constraint subgraphs are often shared across plan skeletons.
To avoid duplicated sampling, we cache the outputs of each sampler, allowing us to reuse samples whenever we encounter a shared subgraph.
This significantly decreases sampling time by avoiding repeated calls to samplers with the same parameters.
In our running example, any plan skeleton involving picking the \pddl{red} object at its initial placement \(\bm{p_0}\), \pddl{Pick}(\pddl{red}, $g_i$, $\bm{p_0}$, $q_i$), shares the subgraph in Figures~\ref{fig:individual-sampling-networks}a and~\ref{fig:individual-sampling-networks}b, corresponding to the grasp and robot configuration samplers.

\section{Approximating Plan Skeleton Feasibility}
\label{sec:search-over-skeletons}
Now that we have discussed both particle initialization and optimization given a fixed plan skeleton, we turn to the final component of~\gputamp{}: searching over plan skeletons.
A key challenge in sequence-then-satisfy approaches to TAMP is to avoid the \textit{refinement} of plan skeletons with CSPs that are unsolvable, as this results in wasted computational effort.
However, proving whether a CSP with non-linear and non-convex constraints is unsolvable is generally intractable.
In sampling-based TAMP, the typical strategy is to sample up to a maximum budget and backtrack upon failure to explore the next skeleton.
Our strategy in \gputamp{} leverages the thousands of candidate solutions generated during particle initialization to estimate plan skeleton feasibility.
We use this feasibility measure, along with other factors such as plan skeleton length, to determine the order in which we refine plan skeletons (\textsc{PlanHeuristic} in Algorithm~\ref{alg:gpu-tamp}).

Recall that our parallelized samplers in Section~\ref{sec:particle-initialization} attempt to solve constraint subgraphs (Figure~\ref{fig:individual-sampling-networks}).
Proving that a subgraph is feasible requires the sampler to find just one counterexample out of the batch of \(N_b\) particles.
Thus, if any constraints in a plan skeleton have zero satisfying particles, we can say it is likely to be infeasible as we scale \(N_b\).

We use this insight to derive a  heuristic for the feasibility of a plan skeleton \(\pi\) with constraints \(\textbf{con}(\pi)\) and particles \(\mathcal{P}\):
\vspace{-5pt}
\begin{equation}
\label{eq:plan-feasibility}
\begin{split}
    &H(\pi, \mathcal{P}) = \frac{1}{|\textbf{con}(\pi)|} \sum_{c \in \textbf{con}(\pi)} h(\mathcal{P}, c), \\
    \text{where } \ &h(\mathcal{P}, c) =
    \begin{cases}
    \Lambda_\text{penalty} &\text{if } n_{\text{satisfying}} = 0,\\
    n_{\text{satisfying}} &\text{otherwise}.
    \end{cases}
\end{split}
\end{equation}

\(n_{\text{satisfying}} = |J_{c}(\mathcal{P}_{\textbf{param}(c)}) \leq \epsilon_{c}|\) is the number of particles that satisfy constraint \(c\) (see Sec.~\ref{sec:csp-as-optimization}),
\(\Lambda_\text{penalty}\) denotes a large negative penalty applied when no particle in the batch satisfies constraint \(c\).
Intuitively, \(H(\pi, \mathcal{P})\) measures the average feasibility of constraints across the plan skeleton by counting the number of particles satisfying each constraint, and assigning a large penalty to constraints with no satisfying particles.
This heuristic can be quickly evaluated on the GPU.

\textbf{Pruning Plan Skeletons with Failed Subgraphs.} 
Particle initialization is relatively inexpensive when compared to differentiable optimization, but it still requires non-trivial computation per skeleton.
TAMP problems involving many objects may admit hundreds or even thousands of plan skeletons, the majority of which may be infeasible.
We seek to detect plan skeletons that have the same pattern of failure as previously unsuccessfully sampled plan skeletons.

To achieve this, we first define a constraint to be likely unsatisfiable if it has zero satisfying particles after sampling.
We collect such constraints and prune new skeletons from consideration by the task planner if their constraint networks contain the same unsatisfiable constraint subgraph.
For example, in Figure~\ref{fig:stick-button-rollout}, the blue button is out of reach of the Franka, requiring it to use the stick to press the button.
The kinematic constraint for pressing the blue button directly without the stick would always have zero satisfying particles.
Thus, any plan skeleton that attempts to press the blue button directly can be pruned, as it is infeasible.

Because it is generally intractable to prove constraint unsatisfiability, we periodically re-initialize particles for the parameters involved in the failed constraint subgraphs.
If we find a counterexample, i.e., at least one particle that satisfies the constraint, we add all corresponding pruned skeletons back to the search queue, as the constraint has been proven feasible.

\section{Experimental Evaluation.}
\label{sec:experiments}

We evaluate \gputamp{} on a range of simulated TAMP problems with varying levels of difficulty. These problems differ in the number of plan skeletons that are feasible, and the volume of the solution space due to the complex interaction between constraints.
We conduct ablation studies on \gputamp{} to analyze the impact of particle batch size, tuning of the cost weights \(\lambda\), and subgraph caching.
We consider the following baselines, which emulate existing serial sampling-based~\cite{garrett2018sampling} and optimization-based~\cite{toussaint2015logic} planners when run with a single particle (\(N_b = 1\)):
\paragraph{\textsc{Sampling}}
Particles are continuously resampled without optimization via \proc{OptimizeParticles}.
\paragraph{\textsc{Optimization}}
Particles are initialized uniformly within bounds (e.g., joint limits) without sampling. %

\begin{figure}[t]
    \centering
    \begin{subfigure}[t]{0.49\linewidth}
        \centering
        \includegraphics[width=\textwidth]{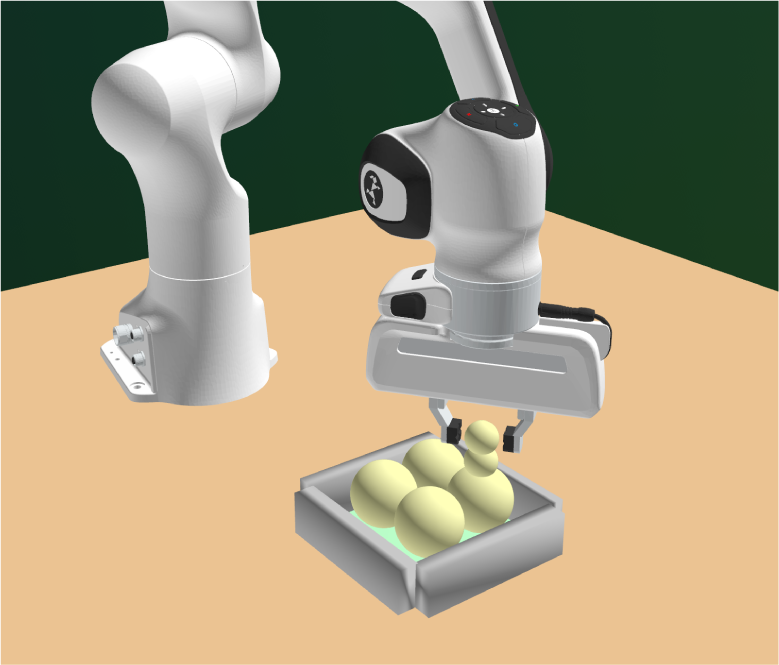}
        \caption{Single Object Packing}
        \label{fig:tetris-1}
    \end{subfigure}
    \hfill
    \begin{subfigure}[t]{0.49\linewidth}
        \centering
        \includegraphics[width=\textwidth]{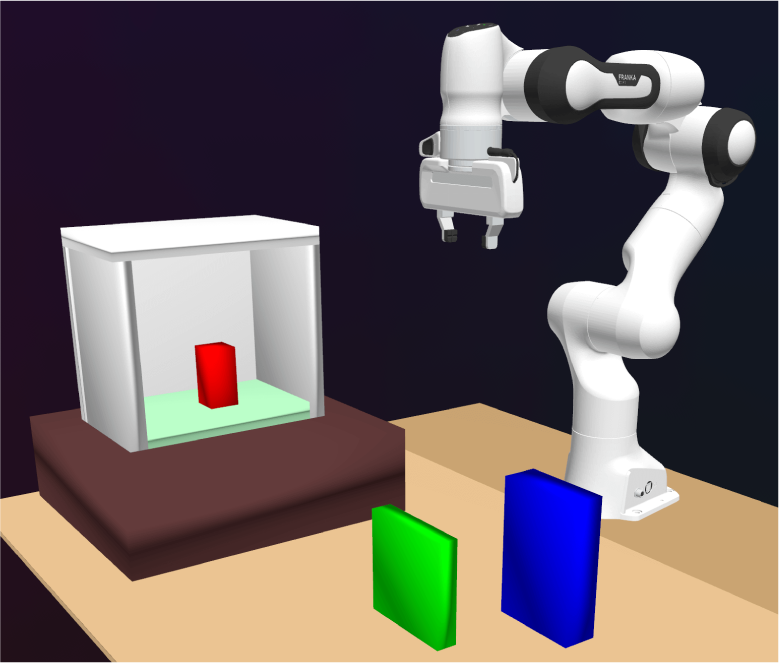}
        \caption{Bookshelf with Obstacle}
        \label{fig:book-shelf}
    \end{subfigure}
    \caption{
        \textbf{TAMP Problems with Obstructions.} 
        (a) Requires packing a square-shaped block. %
        (b) Requires packing the blue and green books into a shelf with a red obstacle. 
    }
    \label{fig:tamp-obstruction-domains}
    \vspace{-10px}
\end{figure}

\textbf{Experimental Setup.}
For each approach, we run at least 10 trials and report the coverage (i.e., success rate) along with the mean and 95\% confidence interval across metrics including the number of satisfying particles and runtime to find a solution.
Grasps are parametrized as top-down 4-DOF or 6-DOF poses, while placements are only 4-DOF poses. Robot configurations correspond to 7-DOF joint positions.
Grasps are always sampled and fixed, while placement poses and robot configurations are optimized.
To evaluate the generality of \gputamp{}, we use the same cost weights across all problems and the same learning rates for Adam.
The primary hyperparameter we vary is the particle batch size \(N_b\), in order to investigate how increased parallelism affects performance.

We defer motion generation until after solving for placements and robot configurations.
While \gputamp{} supports directly jointly optimizing collision-free trajectories (Figure~\ref{fig:rummy-block-stacking}), we found that in practice it is more computationally efficient to optimize for collision-free start and end configurations and defer full motion planning until after optimization.
Specifically, we iterate through satisfying particles until we find one that admits full motions in a semi-hierarchical fashion.
We use the GPU-based motion planner within cuRobo~\cite{sundaralingam2023curobo}.
The timing information we present does not include this motion planning time, which requires only around a few hundred milliseconds for each trajectory parameter and almost always succeeds for our distribution of problems.
Additional experimental details and full results can be found in Appendix~\ref{app:experiments}.

\begin{figure}[t]
\centering
\begin{footnotesize}{
\setlength{\tabcolsep}{4pt}
\begin{tabular}{lrrrr}
\toprule
\textbf{Approach} & $N_b$ & \textbf{Coverage} & \textbf{\#Satisfying} & \textbf{Sol. Time (s)} \\
\midrule
\multirow{4}{*}{\textsc{Sampling}} & 1 & 30/30 & 11.33 ± 1.35 & 0.648 ± 0.171 \\
 & 64 & 30/30 & 594.70 ± 9.82 & \textbf{\hl{0.096 ± 0.011}} \\
 & 256 & 30/30 & 1419.20 ± 12.97 & 0.100 ± 0.002 \\
 & 1024 & 30/30 & 1839.07 ± 14.45 & 0.163 ± 0.001 \\
\midrule
\multirow{4}{*}{\textsc{Optimization}} & 1 & 4/30 & 1.00 ± 0.00 & 3.096 ± 1.914 \\
 & 64 & 30/30 & 5.63 ± 0.95 & 2.383 ± 0.192 \\
 & 256 & 30/30 & 22.20 ± 1.82 & 1.857 ± 0.113 \\
 & 1024 & 30/30 & 91.17 ± 3.93 & \textbf{1.394 ± 0.082} \\
\midrule
\multirow{4}{*}{\gputamp{}} & 1 & 26/30 & 0.85 ± 0.15 & 0.530 ± 0.101 \\
 & 64 & 30/30 & 46.67 ± 1.37 & 0.138 ± 0.041 \\
 & 256 & 30/30 & 192.67 ± 2.31 & \textbf{0.099 ± 0.002} \\
 & 1024 & 30/30 & 770.97 ± 5.99 & 0.163 ± 0.002 \\
\bottomrule
\end{tabular}

}
\end{footnotesize}
\captionof{table}{
\textbf{Results on Single Object Packing.}
We ablate the particle batch size \(N_b\), where \(N_b = 1\) is representative of serial approaches~\cite{garrett2018sampling,toussaint2015logic}. The \#Satisfying metric measures the number of satisfying particles. The best solution time for each approach is bolded, and the overall best is \hl{highlighted}.
}
\label{tab:tetris-1}
\vspace{-10px}
\end{figure}

\subsection{Solving TAMP Problems with Obstruction}

\textbf{Single Object Packing} (Figure~\ref{fig:tetris-1}).
The TAMP planner must find placement poses for the object that do not collide with the grey walls.
Each method is given a 5 second budget for resampling or optimization.
Table~\ref{tab:tetris-1} shows that all methods successfully find solutions in at least some trials.
Due to the short horizon of this problem, \textsc{Sampling} alone with a small batch size is sufficient to solve it.
However, \textsc{Optimization} with uniform random initialization performs poorly.
Although increasing the batch size improves its performance, it remains \(14 \times\) slower than the other approaches to find satisfying solutions.
This highlights the importance of particle initialization via conditional sampling to start near the solution manifold.

\textbf{Bookshelf Problem} (Figure~\ref{fig:book-shelf}). 
The robot can choose either to: 1) pack the books directly into the shelf without removing the red obstacle, or 2) first remove the obstacle before packing the books. The CSP induced by the first strategy is significantly more constrained, requiring increased sampling and optimization time to find a satisfying solution.
We sample 6-DOF grasps in this problem, as they are required to ensure collision-free placements of the books.

Our results are presented in Table~\ref{tab:book-shelf}. We observe that the plan feasibility heuristic (Eq.~\ref{eq:plan-feasibility}) prioritizes skeletons that first move the obstacle out of the way, as they result in more satisfying solutions.
Both variants of \gputamp{} are over 45\% faster than \textsc{Sampling} to find a solution and achieve full coverage with much smaller batch sizes.
Subgraph caching improves the runtime of \gputamp{} by 28\%, as it speeds up particle initialization over skeletons.
However, it can suffer from local minima with smaller batch sizes.

\begin{figure}[t]
\centering
\begin{footnotesize}{
\setlength{\tabcolsep}{3pt}
\begin{tabular}{lrrrrrr}
\toprule
\textbf{Approach} & $N_b$ & \textbf{Coverage} & \textbf{\#Opt. Plans} & \textbf{Sol. Time (s)} \\
\midrule
\multirow{4}{*}{\parbox{1.35cm}{\textsc{Sampling}}} & 64 & 7/30 & 1.57 ± 1.05 & \textbf{3.95 ± 3.65} \\
 & 256 & 24/30 & 1.54 ± 0.37 & 5.83 ± 2.29 \\
 & 512 & 28/30 & 1.25 ± 0.17 & 7.22 ± 1.72 \\
 & 2048 & 30/30 & 1.27 ± 0.26 & 11.59 ± 3.15 \\
\midrule
\multirow{4}{*}{\parbox{1.35cm}{\gputamp{}}} & 64 & 30/30 & 1.80 ± 0.65 & 4.09 ± 1.29 \\
 & 256 & 30/30 & 1.00 ± 0.00 & \textbf{2.11 ± 0.17} \\
 & 512 & 30/30 & 1.03 ± 0.07 & 2.68 ± 0.19 \\
 & 2048 & 30/30 & 1.00 ± 0.00 & 6.59 ± 0.72 \\
\midrule
\multirow{4}{*}{\parbox{1.35cm}{\gputamp{} w. subgraph caching}} & 64 & 29/30 & 1.93 ± 0.87 & 4.61 ± 2.67 \\
 & 256 & 30/30 & 1.03 ± 0.07 & 1.61 ± 0.39 \\
 & 512 & 30/30 & 1.00 ± 0.00 & \textbf{\hl{1.52 ± 0.15}} \\
 & 2048 & 30/30 & 1.00 ± 0.00 & 2.15 ± 0.27 \\
\bottomrule
\end{tabular}

}
\end{footnotesize}
\captionof{table}{
\textbf{Results on Bookshelf.} \#Opt. Plans metric measures the number of plan skeletons that were optimized or resampled before a solution was found (Stage 2 of Algorithm~\ref{alg:gpu-tamp}).}
\label{tab:book-shelf}
\end{figure}

\subsection{Optimizing Goal Costs}
\label{sec:opt-goal-costs}

\begin{figure}[t]
\centering
\begin{footnotesize}
\begin{tabular}{lrrrr}
\toprule
\textbf{Approach} & $N_b$ & \textbf{\#Satisfying} & \textbf{Best Cost} \\
\midrule
\multirow{4}{*}{\parbox{1.25cm}{\textsc{Sampling}}} & 256 & 3380.70 ± 39.54 & 63.25 ± 2.19 \\
 & 512 & 4249.40 ± 68.54 & 62.66 ± 2.76 \\
 & 2048 & 4795.00 ± 39.19 & \textbf{60.00 ± 3.35} \\
 & 4096 & 5136.10 ± 49.83 & 60.14 ± 3.51 \\
\midrule
\multirow{4}{*}{\parbox{1.25cm}{\gputamp{} w.o. goal cost}} & 256 & 122.00 ± 4.10 & 68.97 ± 4.99 \\
 & 512 & 269.40 ± 11.39 & 63.37 ± 2.48 \\
 & 2048 & 1092.80 ± 54.68 & 61.25 ± 1.58 \\
 & 4096 & 1936.20 ± 54.06 & \textbf{58.34 ± 1.86} \\
\midrule
\multirow{4}{*}{\parbox{1.25cm}{\gputamp{} $\lambda$ = 0.025}} & 256 & 23.90 ± 2.49 & 44.70 ± 0.55 \\
 & 512 & 48.40 ± 4.39 & \textbf{\hl{44.57 ± 0.53}} \\
 & 2048 & 125.60 ± 11.86 & 45.21 ± 0.65 \\
 & 4096 & 331.80 ± 26.76 & 46.90 ± 0.72 \\
\bottomrule
\end{tabular}

\captionof{table}{
\textbf{Optimizing Goal Costs.}
We minimize the distance between four objects and compare the best particle cost.
\(\lambda\) is the weight applied to the goal cost during optimization.
}
\label{tab:obstacle-blocks-soft}
\vspace{-12px}
\end{footnotesize}
\end{figure}

We consider minimizing the goal distance between four objects in a goal region (Figure~\ref{fig:obstacle-blocks-soft}).
We compare \textsc{Sampling} against two versions of \gputamp{}: one that explicitly optimizes the goal cost and one without.
For each approach in Table~\ref{tab:obstacle-blocks-soft}, we allocate 10 seconds for optimization or resampling.
Our results demonstrate the clear benefit of combining sampling with differentiable optimization for substantially reducing the goal cost.
\textsc{Sampling} struggles to find good solutions compared to \gputamp{} as shown in Figure~\ref{fig:obstacle-blocks-soft}.

\subsection{Solving Highly-Constrained Problems}
\label{sec:tetris}

\begin{figure}[t]
\centering
\begin{footnotesize}
\setlength{\tabcolsep}{3pt}
\begin{tabular}{lrrrr}
\toprule
\textbf{Approach} & $N_b$ & \textbf{Coverage} & \textbf{\#Satisfying} & \textbf{Sol. Time (s)} \\
\midrule
\multirow{1}{*}{\parbox{1.35cm}{\textsc{Sampling}}} & 4096 & 0/50 & -- & -- \\
\midrule
\multirow{4}{*}{\parbox{1.35cm}{\gputamp{}}} & 512 & 30/50 & 0.70 ± 0.30 & \textbf{11.51 ± 2.72} \\
 & 1024 & 34/50 & 1.18 ± 0.33 & 11.55 ± 2.04 \\
 & 2048 & 47/50 & 1.79 ± 0.36 & 11.82 ± 2.69 \\
 & 4096 & 50/50 & 4.10 ± 0.71 & 12.21 ± 2.34 \\
\midrule
\multirow{4}{*}{\parbox{1.35cm}{\gputamp{} Tuned}} & 512 & 48/50 & 1.75 ± 0.34 & 7.13 ± 1.51 \\
 & 1024 & 50/50 & 3.00 ± 0.48 & \textbf{\hl{5.38 ± 0.63}} \\
 & 2048 & 50/50 & 6.54 ± 0.90 & 5.63 ± 0.58 \\
 & 4096 & 50/50 & 11.96 ± 1.14 & 7.99 ± 0.58 \\
\bottomrule
\end{tabular}

\end{footnotesize}
\captionof{table}{
\textbf{Results on Tetris 5 blocks.} 
In \gputamp{} Tuned, we tune the cost weights automatically using Optuna~\cite{akiba2019optuna}.
}
\label{tab:tetris-5-blocks}
\vspace{-12px}
\end{figure}

The objective in the \textit{Tetris} domain is to pack 5 blocks with non-convex shapes into a tight goal region (Figure~\ref{fig:teaser}).
This task requires reasoning about precise positions and orientations,
as the shapes will only fit if they are arranged in particular configuration modes.
Each particle is 90-dimensional (ten 7-DOF arm configurations and five 4-DOF placements), highlighting the complexity of the problem.

\textbf{Hyperparameter Tuning.}
We conduct an ablation of \gputamp{} where we automatically tune the weight \(\lambda_c\) for each cost \(c\) (Eq.~\ref{eq:particle-cost}) using Optuna~\cite{akiba2019optuna}, an off-the-shelf hyperparameter optimization framework, with the number of satisfying particles as the target objective.
To reduce the dimensionality of the tuning space, we group costs by type and assign a shared weight to each group (e.g., all kinematic position error terms share the same weight).
We tune the cost weights over 220 trials, and provide further details in Appendix~\ref{app:tetris-results}.

\textbf{Results.} We present our results in Table~\ref{tab:tetris-5-blocks}.
\textsc{Sampling} fails completely, while both tuned and untuned variants of \gputamp{} successfully find solutions within seconds.
Only 0.3\% of the optimized particles are satisfying, underscoring the advantage of sampling and optimizing thousands of candidate solutions in parallel.
Larger particle batch sizes \(N_b\) consistently improve the success rate, the primary planning metric, especially on highly challenging problems like Tetris.
Appendix Figure~\ref{fig:runtime-tetris-5} shows that the \gputamp{}'s runtime remains roughly constant up to a batch size of 512, then scales linearly beyond that.

Automated tuning of the cost weights more than halves the time required to find a solution and nearly triples the number of satisfying particles.
The tuned weights generalize to a 3 block variant of Tetris, achieving a similar solve time to untuned \gputamp{} while increasing the number of satisfying particles (Table~\ref{tab:tetris-3-blocks}).
However, they can perform worse than untuned \gputamp{} on other problems due to overfitting (Table~\ref{tab:bookshelf-obstacle-full-results}).

\subsection{Efficiently Searching over Plan Skeletons}

\begin{figure}[t]
    \includegraphics[width=\linewidth]{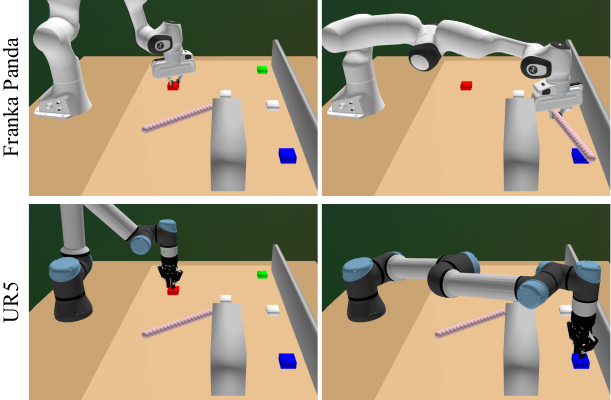}
    \caption{\textbf{Cross-Embodiment Generalization in Stick Button.}
    The blue button is beyond the reach of the Franka, requiring it to use the stick as a tool.
    It plans to grasp near the stick's end to ensure it can press the blue button without colliding into the walls.
    In contrast, the UR5 can directly reach the button.
    }
    \label{fig:stick-button-rollout}
    \vspace{-5px}
\end{figure}

In \text{Stick Button}, the robot must press the red, green, and blue buttons using either its fingers or a stick as a tool (Figure~\ref{fig:stick-button-rollout}).
Due to the kinematic limitations of the Franka, the robot must use the stick to press the blue and green buttons, as they are out of direct reach, demonstrating \gputamp{}'s ability to plan with non-prehensile actions.
This results in a large number of plan skeletons that are infeasible or have extraneous actions, such as pressing the blue button with the robot's fingers or pressing the white distractor buttons.
A TAMP planner must quickly identify skeletons likely to be infeasible, and reason that the stick must be grasped near one of its ends to press the blue button without colliding with the gray walls.

Our results in Table~\ref{tab:stick-button} indicate that \gputamp{} autonomously infers that the stick must be used, leveraging the plan feasibility heuristic to avoid optimizing plan skeletons that are likely infeasible.
This heuristic becomes more accurate as we increase the batch size. 
While \textsc{Sampling} reliably solves stick button with 512 particles, it is over \(3\times\) slower than \gputamp{} with subgraph caching.
Subgraph caching substantially speeds up particle initialization and enables quicker transition to the optimization phase of \gputamp{} (Stage 2 in Algorithm~\ref{alg:gpu-tamp}).
In contrast to the Franka, the UR5 can reach all the buttons directly without using the stick (Figure~\ref{fig:stick-button-rollout}).
When deployed on the UR5, \gputamp{} automatically and implicitly infers this difference in reachability.

\begin{figure}[t]
\centering
\begin{footnotesize}{
\setlength{\tabcolsep}{3pt}
\begin{tabular}{lrrrrrr}
\toprule
\textbf{Approach} & $N_b$ & \textbf{Coverage} & \textbf{\#Opt. Plans} & \textbf{Sol. Time (s)} \\
\midrule
\multirow{4}{*}{\parbox{1.4cm}{\textsc{Sampling}}} & 32 & 3/10 & 2.33 ± 2.87 & 5.10 ± 2.97 \\
 & 64 & 5/10 & 2.20 ± 2.69 & 5.15 ± 5.74 \\
 & 512 & 10/10 & 1.00 ± 0.00 & \textbf{4.78 ± 1.16} \\
 & 2048 & 10/10 & 1.00 ± 0.00 & 12.20 ± 3.77 \\
\midrule
\multirow{4}{*}{\parbox{1.4cm}{\gputamp{}}} & 32 & 10/10 & 4.60 ± 6.44 & 4.50 ± 2.31 \\
 & 64 & 10/10 & 1.20 ± 0.30 & \textbf{2.94 ± 0.56} \\
 & 512 & 10/10 & 1.00 ± 0.00 & 4.37 ± 0.55 \\
 & 2048 & 10/10 & 1.00 ± 0.00 & 10.82 ± 2.41 \\
\midrule
\multirow{4}{*}{\parbox{1.4cm}{\gputamp{} w. subgraph caching}} & 32 & 8/10 & 10.00 ± 10.37 & 7.96 ± 3.75 \\
 & 64 & 10/10 & 3.30 ± 1.22 & 8.72 ± 3.33 \\
 & 512 & 10/10 & 1.00 ± 0.00 & \textbf{\hl{1.45 ± 0.05}} \\
 & 2048 & 10/10 & 1.00 ± 0.00 & 2.06 ± 0.04 \\
\bottomrule
\end{tabular}

}
\end{footnotesize}
\captionof{table}{\textbf{Results on Stick Button}
The \#Opt. Plans metric measures the number of plans that were optimized or resampled before a solution was found (Stage 2 of Algorithm~\ref{alg:gpu-tamp}).
}
\label{tab:stick-button}
\vspace{-5px}
\end{figure}

\subsection{Real Robot Experiments}

\begin{figure}[t]
\centering
\includegraphics[width=\linewidth]{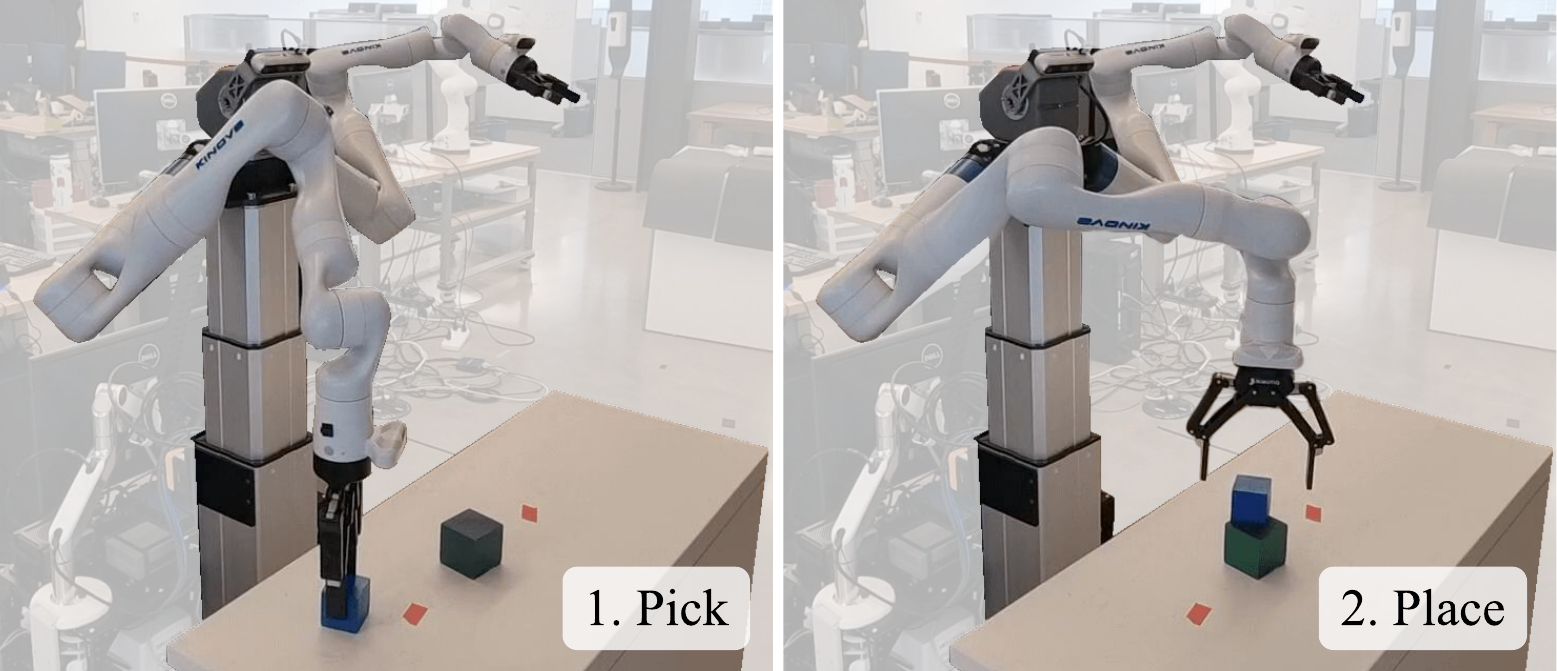}
\caption{\textbf{Real-World Block Stacking} We jointly optimize grasps, placements, and trajectory knot points using \gputamp{}.} %
\label{fig:rummy-block-stacking}
\vspace{-2px}
\end{figure}

\begin{figure}[t]
  \centering
  \includegraphics[width=\linewidth]{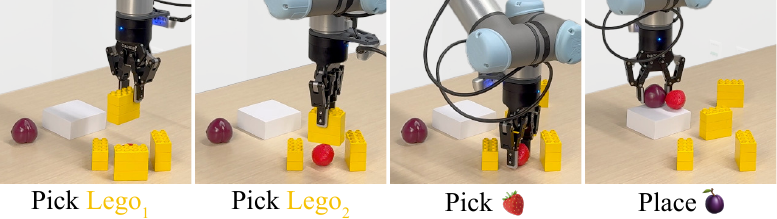}
  \caption{
  \textbf{Packing fruit with obstacles.} The strawberry is obstructed by four Lego blocks, requiring at least two to be moved for a feasible grasp. \gputamp{} autonomously infers this to find a feasible plan skeleton and valid solution.
  }
  \label{fig:fruit-packing-real}
  \vspace{-10px}
\end{figure}

We deploy \gputamp{} on two embodiments: 1) a UR5 arm, %
and 2) a Kinova Gen3 robot. %
We use the open-world perception strategy from~\cite{curtis2022long} to reconstruct the objects and tabletop.
We minimize the goal distance between four objects on a UR5 in Figure~\ref{fig:ur5-min-obj-dist} and achieve a tight packing.
In Figures~\ref{fig:rummy-obstruction} and~\ref{fig:fruit-packing-real}, \gputamp{} autonomously reasons to move obstacles out of the way to achieve the goal on a Kinova Gen3 and UR5, respectively.
Finally, we show that \gputamp{} supports jointly optimizing grasps, placements, and \textit{full trajectories} parametrized as knot points in Figure~\ref{fig:rummy-block-stacking}. Trajectories are initialized by linearly interpolating between the start and end robot configurations.
We highly encourage the reader to refer to our {\color{bluelink}\href{https://cutamp.github.io}{website}} and supplementary video, which showcases the fast planning capabilities of \gputamp{} in the real world.

\section{Limitations}

While cuTAMP supports 6-DOF grasps, we model placement poses with 3-DOF position and yaw (Section~\ref{sec:experiments}).
In the future, we would like to also model varying placement roll and pitch, accounting for stable orientations on approximately planar surfaces.
Our experiments show that \gputamp{}'s optimal configuration is sensitive to the number of particles and cost weights \(\lambda\), though it performs robustly with the default weights across all tested domains. Future work could explore augmented Lagrangians to dynamically scale costs to satisfy constraints~\cite{platt1987constrained,toussaint2014newton}, as well as particle pruning to avoid optimizing particles that converge early due to local minima.

\gputamp{} leverages parallelism during continuous optimization but does not parallelize the search of plan skeletons.
GPU-accelerated task planning~\cite{dal2015cud}, via Boolean Satisfiability (SAT) solvers~\cite{kautz1992planning}, may be able to parallelize all of planning.
Our approach addresses deterministic observable planning. Future work involves leveraging GPU acceleration to reason about stochastic actions and partial observability~\cite{kaelbling2013integrated,garrett2020online}.
Extending cuTAMP to tasks involving more complex non-prehensile and contact-rich manipulation is another important direction.

\section{Conclusion}

We propose \gputamp{}, the first GPU-parallelized TAMP algorithm for robot manipulators.
\gputamp{} uses GPU acceleration first during compositional sampling to generate an initial set of plan parameter particles,
and second, during joint optimization to solve for particles that satisfy plan constraints and minimize plan costs.
We show that increasing the number of particles, particularly beyond a single particle, improves planning runtime and solution quality in challenging, long-horizon manipulation problems.
We deploy \gputamp{} on several robot embodiments, demonstrating its real-world planning capabilities from perception to action in seconds.

\section*{Acknowledgments}

The authors thank Balakumar Sundaralingam for his extensive support with using and debugging cuRobo, and Phillip Isola for providing access to his UR5 robot.
We gratefully acknowledge support from NSF grant 2214177; from AFOSR grant FA9550-22-1-0249; from ONR MURI grants N00014-22-1-2740 and N00014-24-1-2603; from the MIT Quest for Intelligence; and from the Robotics and Artificial Intelligence Institute.

\bibliographystyle{unsrtnat}
\bibliography{references}

\pagebreak

\section*{Appendix}

\setcounter{section}{0}
\setcounter{equation}{0}
\setcounter{figure}{0}
\setcounter{table}{0}

\renewcommand{\thesection}{A\arabic{section}}
\renewcommand{\thefigure}{A.\arabic{figure}}
\renewcommand{\thetable}{A.\arabic{table}}
\renewcommand{\theequation}{A.\arabic{equation}}

\noindent
\textbf{Table of Contents:}
\begin{itemize}
    \item Appendix~\ref{app:problem-formulation}: Expanded problem formulation
    \item Appendix~\ref{appendix:cutamp-prob-comp}: Theoretical analysis
    \item Appendix~\ref{app:configuration}: Experimental configuration
    \item Appendix~\ref{app:experiments}: Expanded experimental results
\end{itemize}

\section{Problem Formulation}
\label{app:problem-formulation}

We expand on our formulation in Section~\ref{sec:problem-formulation} by fully defining types, predicates, constraints, and actions used in our formulation.

\vspace{0.5em}

\noindent \textbf{Types:}
\begin{itemize}
    \item \pddl{conf} - a robot configuration.
    \item \pddl{traj} - a robot trajectory comprised of a sequence of configurations.
    \item \pddl{obj} - a manipulable object.
    \item \pddl{grasp} - an object grasp pose.
    \item \pddl{placement} - an object placement pose.
    \item \pddl{surface} - a placement surface.
    \item \pddl{button} - a button that can be pressed.
    \item \pddl{press} - a pose at which we press a button.
\end{itemize}

\vspace{0.5em}

\noindent \textbf{Predicates:}
\begin{itemize}
    \item \pddl{AtConf}($q$: \pddl{conf}) - the robot is currently at configuration $q$.
    \item \pddl{HandEmpty}() - the robot's hand is currently empty.
    \item \pddl{AtPlacement}($o$: \pddl{obj}, $p$: \pddl{placement}) - object $o$ is currently at placement pose $p$.
    \item \pddl{Holding}($o$: \pddl{obj}, $g$: \pddl{grasp}) - object $o$ is currently grasped with grasp pose $g$.
    \item \pddl{On}($o$: \pddl{obj}, $s$: \pddl{surface}) - object $o$ is placed on top of surface $s$.
    \item \pddl{IsStick}($o$: \pddl{obj}) - object $o$ is a stick we can use to push a button.
    \item \pddl{Pressed}($b$: \pddl{button}) - button $b$ has been pressed.
\end{itemize}

\vspace{0.5em}

\noindent \textbf{Constraints:}
\begin{itemize}
    \item \pddl{Motion}($q_1$: \pddl{conf}, $\tau$: \pddl{traj}, $q_2$: \pddl{conf}) - $\tau$ is a trajectory that connects configurations $q_1$ and $q_2$, and is within the joint limits of the robot.
    \item \pddl{Kin}($q$: \pddl{conf}, $o$: \pddl{obj}, $g$: \pddl{grasp}, $p$: \pddl{placement}) - configuration $q$ satisfies a kinematics constraint with placement pose $p$ when object $o$ is grasped with grasp pose $g$.
    \item \pddl{Grasp}($o$: \pddl{obj}, $g$: \pddl{grasp}) - $g$ is a valid grasp pose for object $o$.
    \item \pddl{StablePlace}($o$: \pddl{obj}, $p$: \pddl{placement}, $s$: \pddl{surface}) - $p$ is a stable placement pose for object $o$ on surface $s$.
    \item \pddl{CFreeTraj}($\tau$: \pddl{traj}) - $\tau$ is a collision-free trajectory with respect to the objects in the world and does not cause robot self-collisions.
    \item \pddl{CFreeHold}($o$: \pddl{obj}, $g$: \pddl{grasp}, $q$: \pddl{conf}) - $g$ is a collision-free grasp for object $o$ using robot configuration $q$.
    \item \pddl{CFreeTrajHold}($o$: \pddl{obj}, $g$: \pddl{grasp}, $\tau$: \pddl{traj}) - trajectory $\tau$ while holding object $o$ with grasp $g$ is collision-free with respect to the objects in the world and does not cause robot self-collisions.
    \item \pddl{CFreePlace}($o$: \pddl{obj}, $p$: \pddl{placement}) - placing object $o$ at placement pose $p$ is collision-free.
    \item \pddl{ValidPress}($b$: \pddl{button}, $p$: \pddl{press}, $q$: \pddl{conf}) - $p$ is a valid press pose for button $b$ and would make contact.
    \item \pddl{ValidStickPress}($o$: obj, $g$: grasp, $p$: press, $b$: button) - $p$ is a valid press pose for button $b$ when using the stick $o$ which is held with grasp pose $p$, i.e., any point along the stick would make contact with the button. 

\end{itemize}

\vspace{0.5em}

\noindent \textbf{Actions:}
\begin{itemize}
    \item \pddl{MoveFree}($q_1$, $q_2$: \pddl{conf}, $\tau$: conf) - move in free space from robot configuration $q_1$ to $q_2$ using trajectory $\tau$.
    \item \pddl{Pick}($o$: obj, $g$: grasp, $p$: placement, $q$: conf - pick object \(o\), which is at placement pose $p$, with grasp \(g\) and corresponding robot configuration \(q\).
    \item \pddl{MoveHolding}($o$: obj, $g$: grasp, $p$: placement, $q$: conf) - move in free space from robot configuration $q_1$ to $q_2$ using trajectory $\tau$ while holding object $o$ with grasp $g$.
    \item \pddl{Place}($o$: obj, $g$: grasp, $p$: placement, $s$: surface, $q$: conf) - place object $o$ which is being held with grasp $g$ on surface $s$ with placement pose $p$, and corresponding robot configuration $q$.
    \item \pddl{PressButton}($b$: button, $p$: press, $q$: conf) - press button $b$ with press pose $p$ at corresponding robot configuration $q$.
    \item \pddl{PressButtonStick}($b$: button, $o$: obj, $g$: grasp, $p$: press, $q$: conf) - press button $b$ at press pose $p$ with the stick $o$, which is being held with grasp $g$, and robot configuration $q$.
\end{itemize}

\begin{figure}[t]
\begin{small}
\begin{lstlisting}
PressButton|$(b: \text{button},\; p: \text{press},\; q: \text{conf})$|
 |\kw{con}:| [Kin|$(q, b, p)$|, ValidPress|$(b, p, q)$|]
 |\kw{pre}:| [HandEmpty|$()$|, AtConf|$(q)$|]
 |\kw{eff}:| [Pressed|$(b)$|]

PressButtonStick|$(b: \text{button},\; o: \text{obj},\; g: \text{grasp},\; p: \text{press},\; q: \text{conf})$|
 |\kw{con}:| [Kin|$(q, o, g, p)$|, ValidStickPress|$(o, g, p, b)$|]
 |\kw{pre}:| [IsStick|$(o)$|, Holding|$(o, g)$|, AtConf|$(q)$|]
 |\kw{eff}:| [Pressed|$(b)$|]
\end{lstlisting}%

\end{small}
\caption{
\textbf{Additional Parametrized Actions.} We list the additional actions required for the Stick Button domain in Figure~\ref{fig:stick-button-rollout}.}
\label{lst:additional-actions}
\end{figure}

\section{Theoretical Analysis of \gputamp{}}
\label{appendix:cutamp-prob-comp}

In this section, we prove that a simplified version of \gputamp{} is probabilistically complete, which informally means that it will almost surely solve nondegenerate feasible problems.
Thus, \gputamp{} has comparable theoretical properties to prior sampling- and optimization-based TAMP solvers~\cite{garrett2018ffrob,dantam2018incremental,garrett2018sampling,vega2020asymptotically}.
Note that TAMP is actually decidable~\cite{deshpande2020decidability,vega2020task}; however, the decomposition-based algorithms provided to prove decidability are computationally inefficient in practice.

Like in constrained motion planning~\cite{kingston2018sampling}, we make the distinction between equality constraints (e.g. kinematics constraints \pddl{Kin}) and inequality constraints (e.g. collision constraints \pddl{CFreeTraj}).
We assume that the real-valued function $J_{c}(\cdot)$ implicitly defining each equality constraint $c$ is continuously differentiable and its Jacobian has full rank. %
As a result, the set of satisfying values is a lower-dimensional submanifold of the ambient constraint space, and thus the constraint is dimensionality-reducing.
Following prior work~\cite{kingston2018sampling} and usage in practice, we consider equality constraints satisfied if $||J_{c}(\xparam{c})||_2 < \epsilon$ for a small $\epsilon > 0$, namely values close to this submanifold are deemed satisfying, comprising an open subset of the ambient space.

We begin by defining {\em robust feasibility}, a property of a problem that holds the problem admits a non-degenerate set of solutions.
Let \(\mathbf{X}_\pi = \{\mathbf{x} \in \mathbf{X} \mid \forall c \in \textbf{con}(\pi).\; J_c(\xparam{c}) \leq \epsilon\}\) be the set of parameter values that satisfy the inequality and equality constraints along plan skeleton $\pi$.

\begin{defn} \label{thm:satisfaible}
A plan skeleton \(\pi = (a_1, \dots, a_n)\) with a bounded $d$-dimensional parameter space $\mathbf{X} \subset \mathbb{R}^d$ is {\em robustly satisfiable} if the set of satisfying parameter values $\mathbf{X}_\pi \subseteq \mathbf{X}$ has positive measure.
\end{defn}

\begin{defn} \label{thm:feasible}
A TAMP problem is \(\Pi = \langle \mathcal{A}, s_0,  S_*\rangle\) {\em robustly feasible} if there exists a correct and robustly satisfiable plan skeleton \(\pi = (a_1, \dots, a_n) \in \mathcal{A}^\infty\) that starting from initial state $s_0$ that reaches a goal state \(s_* \in S_*\).
\end{defn}

Finally, we define {\em probabilistic completeness} and prove \gputamp{} is probabilistically complete.

\begin{defn} \label{thm:complete}
A TAMP algorithm is {\em probabilistically complete} if for any robustly feasible problem $\Pi$, it returns a correct solution in a finite amount of time with probability one.
\end{defn}

\begin{thm} \label{app:thm:completeness}
\gputamp{} is probabilistically complete.
\begin{proof}
We analyze a simplified version of \gputamp{} that initializes parameter particles independently and randomly with positive probability density across $\mathbf{X}$ and omits the plan skeleton feasibility heuristic (Section~\ref{sec:search-over-skeletons}).
\gputamp{} performs a backtracking plan-space best-first (e.g.,  breadth-first) search that enumerates the countable set of plan skeletons (Section~\ref{sec:overview}).
For a plan skeleton $\pi$ that is robustly satisfiable, each random initialization has positive probability of sampling a satisfying particle $\mathbf{x} \in \mathbf{X}_\pi$ because $\mathbf{X}_\pi$ has positive measure and \gputamp{} samples with positive probability density over $\mathbf{X}_\pi$.
Thus, the probability that \gputamp{} produces a satisfying particle when randomly restarting the initialization arbitrarily many times is one.
Because \gputamp{} continuously revisits and resamples all identified plan skeletons, which can alternatively be viewed as creating a separate thread per plan skeleton, once identified, a robustly satisfiable plan skeleton will be randomly restarted arbitrarily many times and thus a solution will be identified with probability one.
\end{proof}
\end{thm}

\section{Experimental Configuration} \label{app:configuration}
We evaluate on an NVIDIA RTX 5880 GPU with 14,080 CUDA cores and 48 GB of VRAM. For comparison, a RTX 4090 has 16,384 CUDA cores and 24 GB of VRAM.
Our current implementation of \gputamp{} saturates the CUDA cores with larger batch sizes and uses around 2.5GB and 4.5GB of GPU memory for batch sizes of 1024 and 2048, respectively.

\textbf{Collision Representation.} We use the primitive collision checker from cuRobo~\cite{sundaralingam2023curobo}, which approximates static objects in the world as oriented bounding boxes. %
The robot and movable objects are approximated using collision spheres.

\textbf{Cost Weights.} We use the same set of cost weights \(\lambda\) across all experiments, apart from where we explicitly tune the weights in Tetris with 5 blocks. All weights for constraints are set to 1.0 except for kinematic rotation error (5.0) and stable placements (2.0).

\textbf{Constraint Tolerances.} As exactly satisfying an equality constraint of 0.0 is hard with numerical optimization, we usually set small tolerances. These are the values we used across all experiments. All satisfying particles satisfy all these tolerances.

\begin{itemize}
    \item Collisions with objects in the world: 1 millimeter (mm).
    \item Robot Self-Collisions: 0 penetration distance (meters).
    \item Kinematics: 5mm translational error, 0.05 radians rotational error.
    \item Robot Joint Limits: 0.
    \item Stable Placement: 1 centimeter for checking support on a surface, 1mm for checking object contained within a surface.
\end{itemize}

\textbf{Real UR5 Configuration.} Grasps are sampled and fixed, while placement poses and robot configurations are optimized. Once either a maximum optimization timeout or a set number of satisfying particles is reached, we generate motions using cuRobo.
In the supplementary videos, we interleave the generation and execution of these motions on the UR5.
A large number of failures in real-world trials can be attributed to inaccuracies in our perception system and a noisy depth sensor (Intel RealSense D435). This results in objects being either under- or over-approximated.

\section{Full Experimental Results}
\label{app:experiments}

\subsection{Single Object Packing}

We give a maximum resampling or optimization time of 5 seconds after initialization. The full results are in Table~\ref{tab:tetris-1-full-results}. Since \gputamp{} uses the same particle initialization phase as \textsc{Sampling}, the results are near indistinguishable outside of noise due to randomness.

\begin{figure}[t]
\centering
\resizebox{\linewidth}{!}{%
\begin{tabular}{lrrrr}
\toprule
\textbf{Approach} & $N_b$ & \textbf{Coverage} & \textbf{\#Satisfying} & \textbf{Sol. Time (s)} \\
\midrule
\multirow{13}{*}{\textsc{Sampling}} & 1 & 30/30 & 11.33 ± 1.35 & 0.648 ± 0.171 \\
 & 2 & 30/30 & 21.40 ± 1.40 & 0.383 ± 0.071 \\
 & 4 & 30/30 & 44.47 ± 2.40 & 0.244 ± 0.046 \\
 & 8 & 30/30 & 89.00 ± 3.54 & 0.205 ± 0.033 \\
 & 16 & 30/30 & 170.20 ± 4.82 & 0.178 ± 0.027 \\
 & 32 & 30/30 & 321.23 ± 4.82 & 0.128 ± 0.024 \\
 & 64 & 30/30 & 594.70 ± 9.82 & \textbf{\hl{0.096 ± 0.011}} \\
 & 128 & 30/30 & 1002.17 ± 10.13 & \textbf{\hl{0.096 ± 0.008}} \\
 & 256 & 30/30 & 1419.20 ± 12.97 & 0.100 ± 0.002 \\
 & 512 & 30/30 & 1801.27 ± 19.32 & 0.115 ± 0.002 \\
 & 1024 & 30/30 & 1839.07 ± 14.45 & 0.163 ± 0.001 \\
 & 2048 & 30/30 & 2011.67 ± 18.43 & 0.237 ± 0.002 \\
 & 4096 & 30/30 & 2077.30 ± 15.63 & 0.394 ± 0.001 \\
\midrule
\multirow{13}{*}{\textsc{Optimization}} & 1 & 4/30 & 1.00 ± 0.00 & 3.096 ± 1.914 \\
 & 2 & 4/30 & 1.00 ± 0.00 & 3.698 ± 0.849 \\
 & 4 & 6/30 & 1.00 ± 0.00 & 3.585 ± 0.664 \\
 & 8 & 15/30 & 1.33 ± 0.40 & 3.568 ± 0.482 \\
 & 16 & 25/30 & 1.96 ± 0.48 & 3.182 ± 0.419 \\
 & 32 & 27/30 & 2.70 ± 0.59 & 2.838 ± 0.323 \\
 & 64 & 30/30 & 5.63 ± 0.95 & 2.383 ± 0.192 \\
 & 128 & 30/30 & 10.77 ± 1.48 & 2.180 ± 0.162 \\
 & 256 & 30/30 & 22.20 ± 1.82 & 1.857 ± 0.113 \\
 & 512 & 30/30 & 41.50 ± 3.33 & 1.670 ± 0.125 \\
 & 1024 & 30/30 & 91.17 ± 3.93 & 1.394 ± 0.082 \\
 & 2048 & 30/30 & 177.20 ± 6.33 & \textbf{1.270 ± 0.080} \\
 & 4096 & 30/30 & 331.80 ± 10.06 & 1.279 ± 0.068 \\
\midrule
\multirow{13}{*}{\gputamp{}} & 1 & 26/30 & 0.85 ± 0.15 & 0.530 ± 0.101 \\
 & 2 & 30/30 & 1.67 ± 0.20 & 0.471 ± 0.041 \\
 & 4 & 30/30 & 3.03 ± 0.29 & 0.409 ± 0.044 \\
 & 8 & 30/30 & 6.43 ± 0.40 & 0.304 ± 0.059 \\
 & 16 & 30/30 & 11.83 ± 0.80 & 0.279 ± 0.057 \\
 & 32 & 30/30 & 23.93 ± 1.06 & 0.135 ± 0.041 \\
 & 64 & 30/30 & 46.67 ± 1.37 & 0.138 ± 0.041 \\
 & 128 & 30/30 & 96.93 ± 2.00 & \textbf{0.099 ± 0.017} \\
 & 256 & 30/30 & 192.67 ± 2.31 & \textbf{0.099 ± 0.002} \\
 & 512 & 30/30 & 382.70 ± 3.05 & 0.116 ± 0.002 \\
 & 1024 & 30/30 & 770.97 ± 5.99 & 0.163 ± 0.002 \\
 & 2048 & 30/30 & 1533.97 ± 6.24 & 0.238 ± 0.001 \\
 & 4096 & 30/30 & 3075.97 ± 10.26 & 0.394 ± 0.001 \\
\bottomrule
\end{tabular}

}
\captionof{table}{\textbf{Full Results on Single Object Packing.} This problem (Figure~\ref{fig:tetris-1}) is easy enough to solve via sampling alone. 
\textsc{Optimization} is not robust on small batch sizes, as uniform random initializations are far from solution manifold. 
}
\label{tab:tetris-1-full-results}
\end{figure}

\subsection{Bookshelf}

We provide \gputamp{} an optimization budget of 1000 steps per plan skeleton.
For the parallelized \textsc{Sampling} baseline, we set a maximum of 50 resampling steps per skeleton.
We present the full results in Table~\ref{tab:bookshelf-obstacle-full-results}.

\begin{figure}[t]
\centering
\begin{small}
\begin{tabular}{lrrrrrr}
\toprule
\textbf{Approach} & $N_b$ & \textbf{Coverage} & \textbf{\#Opt. Plans} & \textbf{Sol. Time (s)} \\
\midrule
\multirow{10}{*}{\parbox{1.55cm}{\textsc{Sampling}}} & 8 & 3/30 & 5.33 ± 10.04 & 5.90 ± 17.24 \\
 & 16 & 2/30 & 1.50 ± 6.35 & \textbf{3.77 ± 11.22} \\
 & 32 & 4/30 & 1.75 ± 0.80 & 4.26 ± 2.20 \\
 & 64 & 7/30 & 1.57 ± 1.05 & 3.95 ± 3.65 \\
 & 128 & 15/30 & 1.87 ± 0.81 & 6.32 ± 2.80 \\
 & 256 & 24/30 & 1.54 ± 0.37 & 5.83 ± 2.29 \\
 & 512 & 28/30 & 1.25 ± 0.17 & 7.22 ± 1.72 \\
 & 1024 & 29/30 & 1.31 ± 0.35 & 9.03 ± 2.41 \\
 & 2048 & 30/30 & 1.27 ± 0.26 & 11.59 ± 3.15 \\
 & 4096 & 30/30 & 1.53 ± 0.36 & 12.95 ± 2.16 \\
\midrule
\multirow{10}{*}{\parbox{1.55cm}{\gputamp{}}} & 8 & 23/30 & 6.65 ± 1.70 & 20.00 ± 8.93 \\
 & 16 & 28/30 & 4.75 ± 1.59 & 16.01 ± 6.81 \\
 & 32 & 30/30 & 4.80 ± 1.58 & 8.51 ± 4.11 \\
 & 64 & 30/30 & 1.80 ± 0.65 & 4.09 ± 1.29 \\
 & 128 & 30/30 & 1.13 ± 0.13 & 2.46 ± 0.40 \\
 & 256 & 30/30 & 1.00 ± 0.00 & \textbf{2.11 ± 0.17} \\
 & 512 & 30/30 & 1.03 ± 0.07 & 2.68 ± 0.19 \\
 & 1024 & 30/30 & 1.00 ± 0.00 & 4.22 ± 0.22 \\
 & 2048 & 30/30 & 1.00 ± 0.00 & 6.59 ± 0.72 \\
 & 4096 & 30/30 & 1.00 ± 0.00 & 11.01 ± 1.34 \\
\midrule
\multirow{10}{*}{\parbox{1.55cm}{\gputamp{} w. subgraph caching}} & 8 & 6/30 & 4.50 ± 4.49 & 5.52 ± 3.59 \\
 & 16 & 17/30 & 5.82 ± 1.93 & 11.75 ± 8.83 \\
 & 32 & 27/30 & 4.19 ± 1.66 & 6.74 ± 4.73 \\
 & 64 & 29/30 & 1.93 ± 0.87 & 4.61 ± 2.67 \\
 & 128 & 30/30 & 1.90 ± 0.88 & 2.37 ± 0.59 \\
 & 256 & 30/30 & 1.03 ± 0.07 & 1.61 ± 0.39 \\
 & 512 & 30/30 & 1.00 ± 0.00 & \textbf{\hl{1.52 ± 0.15}} \\
 & 1024 & 30/30 & 1.00 ± 0.00 & 1.73 ± 0.17 \\
 & 2048 & 30/30 & 1.00 ± 0.00 & 2.15 ± 0.27 \\
 & 4096 & 30/30 & 1.00 ± 0.00 & 3.42 ± 0.44 \\
\midrule
\multirow{10}{*}{\parbox{1.55cm}{\gputamp{} w. subgraph caching tuned on Tetris}} & 8 & 11/30 & 5.55 ± 2.63 & 7.72 ± 3.28 \\
 & 16 & 17/30 & 6.24 ± 2.12 & 7.24 ± 3.08 \\
 & 32 & 25/30 & 3.44 ± 1.42 & 10.88 ± 6.03 \\
 & 64 & 29/30 & 2.76 ± 1.21 & 9.21 ± 5.43 \\
 & 128 & 30/30 & 1.73 ± 0.75 & 4.83 ± 4.84 \\
 & 256 & 30/30 & 1.33 ± 0.62 & \textbf{1.87 ± 0.45} \\
 & 512 & 30/30 & 1.00 ± 0.00 & 1.88 ± 0.19 \\
 & 1024 & 30/30 & 1.00 ± 0.00 & 1.98 ± 0.25 \\
 & 2048 & 30/30 & 1.00 ± 0.00 & 2.73 ± 0.28 \\
 & 4096 & 30/30 & 1.00 ± 0.00 & 3.78 ± 0.49 \\
\bottomrule
\end{tabular}

\end{small}
\captionof{table}{
\textbf{Full Results on Bookshelf with Obstacle.} Note that in the last approach, we evaluate the cost weights tuned on Tetris with 5 blocks.%
}
\label{tab:bookshelf-obstacle-full-results}
\end{figure}

\subsection{Minimizing Object Distance} 

Each approach is given a maximum duration of 10 seconds for resampling or optimization.
We ablate the cost weight \(\lambda\) of the goal cost in \gputamp{}, and also consider disabling explicit optimization of the goal cost.

\begin{figure}[h]
    \centering
    \includegraphics[width=\linewidth]{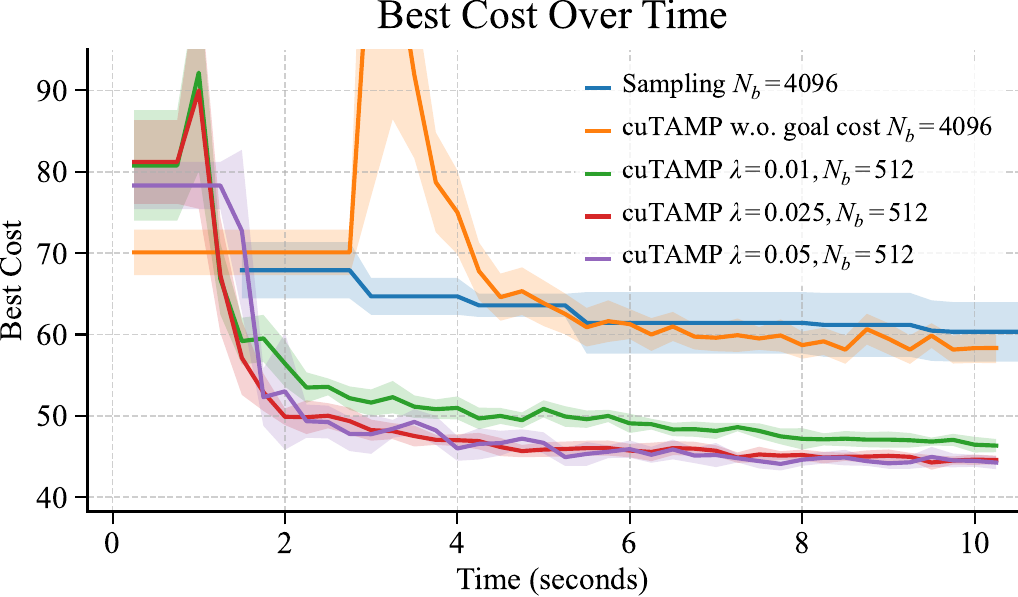 }
    \caption{
    \textbf{Minimizing Object Distance over Time.} We show the cost of the best particle over time as we resample or optimize the particles.
    We plot the best batch size for each approach (refer to Table~\ref{tab:min-obj-dist-full-results}).
    The early spike in the cost for \gputamp{} may be attributed to no warm-up of Adam using a learning rate scheduler, which causes particles to become unsatisfying.
    }
    \label{fig:best-cost-over-time}
\end{figure}

Our full results are in Table~\ref{tab:min-obj-dist-full-results}.
Increasing \(\lambda\) decreases the number of satisfying particles, as the goal cost may conflict with satisfying the hard constraints.
In Figure~\ref{fig:best-cost-over-time}, we plot the best cost over time for the top-performing variant of each approach. \gputamp{} with explicit optimization of the cost significantly outperforms sampling.

\begin{figure}[t]
\centering
\begin{small}
\begin{tabular}{lrrrr}
\toprule
\textbf{Approach} & $N_b$ & \textbf{\#Satisfying} & \textbf{Best Cost} \\
\midrule
\multirow{7}{*}{\parbox{1.5cm}{\textsc{Sampling}}} & 64 & 1477.40 ± 24.18 & 65.54 ± 2.62 \\
 & 128 & 2490.80 ± 36.59 & 65.20 ± 2.70 \\
 & 256 & 3380.70 ± 39.54 & 63.25 ± 2.19 \\
 & 512 & 4249.40 ± 68.54 & 62.66 ± 2.76 \\
 & 1024 & 4330.60 ± 79.62 & 62.49 ± 2.61 \\
 & 2048 & 4795.00 ± 39.19 & \textbf{60.00 ± 3.35} \\
 & 4096 & 5136.10 ± 49.83 & 60.14 ± 3.51 \\
\midrule
\multirow{7}{*}{\parbox{1.2cm}{\gputamp{} w.o. goal cost}} & 64 & 32.10 ± 3.01 & 78.86 ± 8.94 \\
 & 128 & 61.00 ± 5.10 & 72.43 ± 6.41 \\
 & 256 & 122.00 ± 4.10 & 68.97 ± 4.99 \\
 & 512 & 269.40 ± 11.39 & 63.37 ± 2.48 \\
 & 1024 & 563.60 ± 7.82 & 61.36 ± 2.95 \\
 & 2048 & 1092.80 ± 54.68 & 61.25 ± 1.58 \\
 & 4096 & 1936.20 ± 54.06 & \textbf{58.34 ± 1.86} \\
\midrule
\multirow{7}{*}{\parbox{1.3cm}{\gputamp{} $\lambda = 0.1$}} & 64 & 16.40 ± 2.19 & 47.57 ± 0.75 \\
 & 128 & 25.20 ± 3.32 & 47.62 ± 1.24 \\
 & 256 & 57.80 ± 4.84 & 46.67 ± 0.69 \\
 & 512 & 123.40 ± 9.30 & \textbf{46.34 ± 0.81} \\
 & 1024 & 233.20 ± 10.66 & 47.52 ± 0.91 \\
 & 2048 & 566.10 ± 23.32 & 48.60 ± 1.50 \\
 & 4096 & 1506.50 ± 35.92 & 49.53 ± 1.13 \\
\midrule
\multirow{7}{*}{\parbox{1.3cm}{\gputamp{} $\lambda = 0.25$}} & 64 & 8.00 ± 2.55 & 46.60 ± 0.92 \\
 & 128 & 13.20 ± 2.15 & 45.90 ± 0.90 \\
 & 256 & 23.90 ± 2.49 & 44.70 ± 0.55 \\
 & 512 & 48.40 ± 4.39 & \textbf{44.57 ± 0.53} \\
 & 1024 & 79.90 ± 6.81 & 45.31 ± 0.80 \\
 & 2048 & 125.60 ± 11.86 & 45.21 ± 0.65 \\
 & 4096 & 331.80 ± 26.76 & 46.90 ± 0.72 \\
\midrule
\multirow{7}{*}{\parbox{1.3cm}{\gputamp{} $\lambda = 0.5$}} & 64 & 0.90 ± 0.79 & 45.87 ± 1.55 \\
 & 128 & 1.80 ± 0.74 & 45.12 ± 0.84 \\
 & 256 & 4.40 ± 1.40 & 45.17 ± 1.98 \\
 & 512 & 7.50 ± 1.85 & \textbf{\hl{44.26 ± 0.81}} \\
 & 1024 & 10.60 ± 1.94 & 44.35 ± 0.70 \\
 & 2048 & 10.10 ± 2.22 & 45.03 ± 1.05 \\
 & 4096 & 5.30 ± 1.17 & 46.51 ± 1.07 \\
\bottomrule
\end{tabular}

\end{small}
\captionof{table}{
\textbf{Full Results on Minimizing Object Distance.} 
The goal cost is the sum of the distance between the four objects in centimeters. \(\lambda\) is the cost weight we use for the goal cost in the unconstrained optimization problem for \gputamp{}. All approaches achieved full coverage.
}
\label{tab:min-obj-dist-full-results}
\end{figure}

\subsection{Tetris}
\label{app:tetris-results}

Table~\ref{tab:tetris-5-full-results} presents our full results for Tetris with 5 blocks, while Table~\ref{tab:tetris-3-blocks} presents our results for the 3-block variant.
We find that the cost weights tuned on the variant with 5-block generalizes to the 3-block variant.
However, these weights do not transfer to the bookshelf problem (Table~\ref{tab:bookshelf-obstacle-full-results}), where we observe an increase in the time required to find a solution.

\begin{figure*}
  \centering
  \includegraphics[width=.85\linewidth]{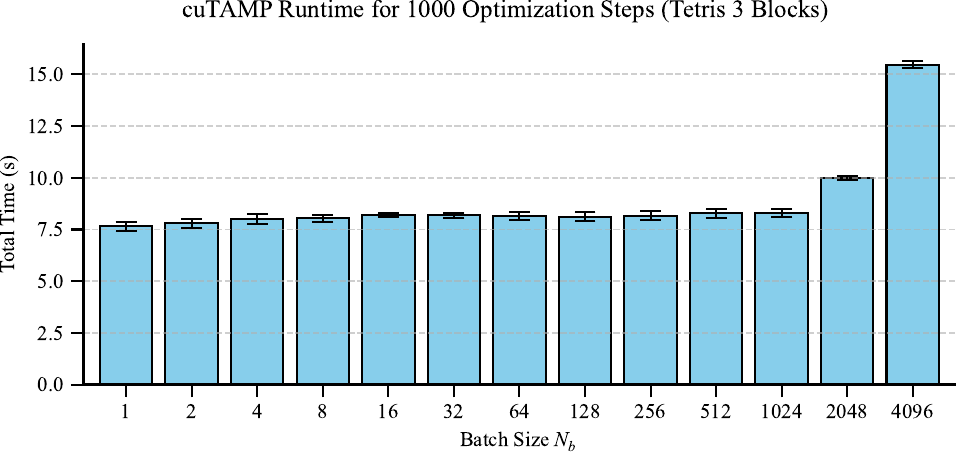}
  \caption{
  \textbf{Runtime of \gputamp{} in Tetris with 3 Blocks.} We show the runtime behavior of \gputamp{}, which includes particle initialization and 1000 differentiable optimization steps, across doubly increasing batch sizes.
    Results are averaged over 10 trials.
    The runtime remains nearly constant for up to 1024 particles, beyond which we observe linear scaling.
    }
  \label{fig:runtime-tetris-3}
\end{figure*}

\begin{figure*}
\centering
\includegraphics[width=.85\linewidth]{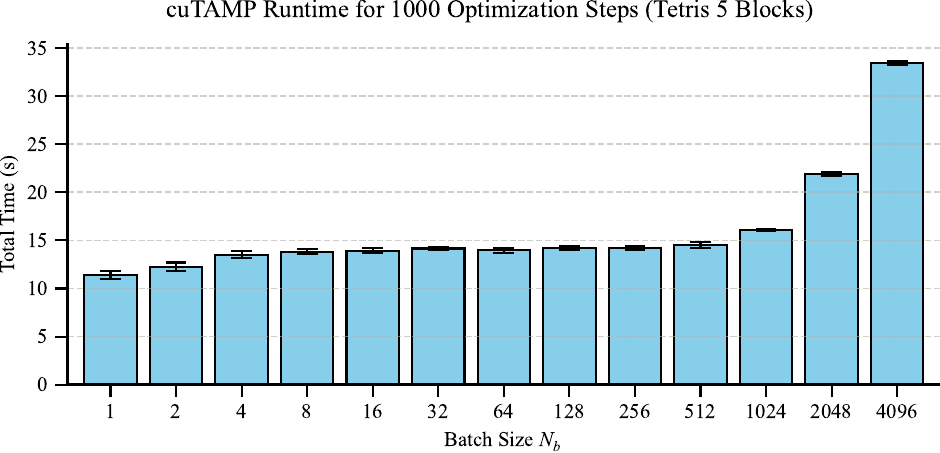}
\caption{
\textbf{Runtime of \gputamp{} in Tetris with 5 Blocks.} We show the runtime behavior of \gputamp{}, which includes particle initialization and 1000 differentiable optimization steps, across doubly increasing batch sizes.
Results are averaged over 10 trials.
The runtime remains nearly constant for up to 512 particles, beyond which we observe linear scaling.
This problem is more difficult and involves more constraints than Tetris with 3 blocks (Figure~\ref{fig:runtime-tetris-3}), hence we observe constant scaling only up to a smaller batch size.
}
\label{fig:runtime-tetris-5}
\end{figure*}

Figures~\ref{fig:runtime-tetris-3} and~\ref{fig:runtime-tetris-5} depict the overall runtime of \gputamp{}, which includes particle initialization and 1000 optimization steps. The runtime stays relatively constant up to 1024 particles in 3-block Tetris, and 512 particles in 5-block Tetris. This demonstrates the benefit of parallelizing on the GPU.

\textbf{Hyperparameter Tuning.}
We tune the cost weights \(\lambda_c\) in the particle cost function (Eq.~\ref{eq:particle-cost}) using Optuna~\cite{akiba2019optuna}, an off-the-shelf hyperparameter optimization framework. To reduce the dimensionality of the tuning space, we group costs by type and assign a shared weight to each group (e.g., all kinematic position error terms share the same weight).  
In each Optuna trial, we run cuTAMP three times on the Tetris problem using the current set of weights, and use the average number of satisfying particles as the objective to maximize.  
We first run 20 trials using a Tree-Structured Parzen Estimator (TPE) sampler~\cite{watanabe2023tree} to efficiently explore the search space and identify promising regions.
We then warm-start a Covariance Matrix Adaptation Evolution Strategy (CMA-ES) sampler~\cite{hansen2016cma} using the TPE trials, and refine the cost weights for an additional 200 trials.

\begin{figure}
\centering
\resizebox{\linewidth}{!}{%
\begin{tabular}{lrrrr}
\toprule
\textbf{Approach} & $N_b$ & \textbf{Coverage} & \textbf{\#Satisfying} & \textbf{Sol. Time (s)} \\
\midrule
\multirow{8}{*}{\parbox{1.6cm}{\textsc{Sampling}}} & 64 & 0/30 & -- & -- \\
 & 128 & 0/30 & -- & -- \\
 & 256 & 0/30 & -- & -- \\
 & 512 & 0/30 & -- & -- \\
 & 1024 & 1/30 & 1.00 ± 0.00 & \textbf{12.31 ± 0.00} \\
 & 2048 & 1/30 & 1.00 ± 0.00 & 18.04 ± 0.00 \\
 & 4096 & 0/30 & -- & -- \\
 & 8192 & 4/30 & 1.00 ± 0.00 & 48.94 ± 23.99 \\
\midrule
\multirow{8}{*}{\parbox{1.35cm}{\gputamp{}}} & 64 & 23/30 & 1.87 ± 0.63 & 1.62 ± 0.44 \\
 & 128 & 30/30 & 2.47 ± 0.66 & 1.57 ± 0.35 \\
 & 256 & 29/30 & 4.93 ± 0.90 & 1.25 ± 0.22 \\
 & 512 & 30/30 & 8.77 ± 1.22 & \textbf{\hl{1.07 ± 0.12}} \\
 & 1024 & 30/30 & 17.33 ± 2.00 & \textbf{\hl{1.07 ± 0.06}} \\
 & 2048 & 30/30 & 36.50 ± 2.80 & 1.30 ± 0.04 \\
 & 4096 & 30/30 & 76.03 ± 4.17 & 1.99 ± 0.06 \\
 & 8192 & 30/30 & 157.47 ± 5.92 & 3.43 ± 0.05 \\
\midrule
\multirow{8}{*}{\parbox{1.35cm}{\gputamp{} Tuned on 5 Blocks}} & 64 & 30/30 & 2.00 ± 0.60 & 1.78 ± 0.49 \\
 & 128 & 29/30 & 3.59 ± 0.82 & 1.71 ± 0.35 \\
 & 256 & 30/30 & 7.23 ± 0.88 & 1.24 ± 0.12 \\
 & 512 & 30/30 & 14.07 ± 1.35 & \textbf{1.09 ± 0.09} \\
 & 1024 & 30/30 & 27.40 ± 2.05 & 1.16 ± 0.05 \\
 & 2048 & 30/30 & 58.47 ± 3.89 & 1.36 ± 0.04 \\
 & 4096 & 30/30 & 117.27 ± 6.70 & 2.05 ± 0.05 \\
 & 8192 & 30/30 & 227.87 ± 12.67 & 3.54 ± 0.06 \\
\bottomrule
\end{tabular}

}
\captionof{table}{
\textbf{Results on Tetris with 3 blocks.} Parallelized \textsc{Sampling} only succeeds due to chance of sampling a valid solution in the limit of more samples. The difference in the time to the first solution is minimal between \gputamp{} and \gputamp{} tuned on Tetris with 5 blocks. However \gputamp{} tuned finds 47\% more satisfying particles on average.}
\label{tab:tetris-3-blocks}
\end{figure}

\begin{figure}
\centering
\resizebox{\linewidth}{!}{%
\begin{tabular}{lrrrr}
\toprule
\textbf{Approach} & $N_b$ & \textbf{Coverage} & \textbf{\#Satisfying} & \textbf{Sol. Time (s)} \\
\midrule
\multirow{8}{*}{\parbox{1.4cm}{\textsc{Sampling}}} & 64 & 0/50 & -- & -- \\
 & 128 & 0/50 & -- & -- \\
 & 256 & 0/50 & -- & -- \\
 & 512 & 0/50 & -- & -- \\
 & 1024 & 0/50 & -- & -- \\
 & 2048 & 0/50 & -- & -- \\
 & 4096 & 0/50 & -- & -- \\
 & 8192 & 0/50 & -- & -- \\
\midrule
\multirow{8}{*}{\parbox{1.4cm}{\gputamp{}}} & 64 & 5/50 & 0.60 ± 0.68 & 15.46 ± 8.31 \\
 & 128 & 6/50 & 0.67 ± 0.54 & 13.37 ± 5.82 \\
 & 256 & 16/50 & 0.69 ± 0.32 & \textbf{9.50 ± 3.53} \\
 & 512 & 30/50 & 0.70 ± 0.30 & 11.51 ± 2.72 \\
 & 1024 & 34/50 & 1.18 ± 0.33 & 11.55 ± 2.04 \\
 & 2048 & 47/50 & 1.79 ± 0.36 & 11.82 ± 2.69 \\
 & 4096 & 50/50 & 4.10 ± 0.71 & 12.21 ± 2.34 \\
 & 8192 & 50/50 & 7.98 ± 0.76 & 14.07 ± 1.48 \\
\midrule
\multirow{8}{*}{\parbox{1.4cm}{\gputamp{} Tuned}} & 64 & 16/50 & 0.50 ± 0.34 & 11.53 ± 4.15 \\
 & 128 & 36/50 & 0.61 ± 0.19 & 11.91 ± 1.77 \\
 & 256 & 44/50 & 1.05 ± 0.26 & 8.54 ± 1.47 \\
 & 512 & 48/50 & 1.75 ± 0.34 & 7.13 ± 1.51 \\
 & 1024 & 50/50 & 3.00 ± 0.48 & \textbf{\hl{5.38 ± 0.63}} \\
 & 2048 & 50/50 & 6.54 ± 0.90 & 5.63 ± 0.58 \\
 & 4096 & 50/50 & 11.96 ± 1.14 & 7.99 ± 0.58 \\
 & 8192 & 50/50 & 23.44 ± 1.66 & 11.29 ± 0.39 \\
\bottomrule
\end{tabular}

}
\captionof{table}{
\textbf{Full Results on Tetris with 5 blocks.} Parallelized \textsc{Sampling} fails across all batch sizes. While untuned \gputamp{} with 256 particles has a low solve time, it has very low coverage. This could be explained by noise in the sampling process, meaning a good initialization has been sampled by chance. Tuned \gputamp{} significantly increases the number of satisfying particles and decreases the time required to find a solution.}
\label{tab:tetris-5-full-results}
\end{figure}

\subsection{Stick Button}

In the \textsc{Sampling} baseline, we limit the maximum resampling steps per plan skeleton to 50.
For \gputamp{}, each skeleton undergoes 500 optimization steps.
Our full results are presented in Table~\ref{tab:stick-button-full-results}.
Enabling subgraph caching in \gputamp{} significantly accelerates the particle initialization over skeletons.

\begin{figure}[t]
\centering
\begin{small}
\begin{tabular}{lrrrrrr}
\toprule
\textbf{Approach} & $N_b$ & \textbf{Coverage} & \textbf{\#Opt. Plans} & \textbf{Sol. Time (s)} \\
\midrule
\multirow{10}{*}{\parbox{1.3cm}{\textsc{Sampling}}} & 8 & 1/10 & 1.00 ± 0.00 & \textbf{1.68 ± 0.00} \\
 & 16 & 4/10 & 6.75 ± 8.46 & 13.48 ± 14.73 \\
 & 32 & 3/10 & 2.33 ± 2.87 & 5.10 ± 2.97 \\
 & 64 & 5/10 & 2.20 ± 2.69 & 5.15 ± 5.74 \\
 & 128 & 8/10 & 1.50 ± 0.63 & 4.73 ± 2.00 \\
 & 256 & 10/10 & 1.20 ± 0.30 & 4.50 ± 1.91 \\
 & 512 & 10/10 & 1.00 ± 0.00 & 4.78 ± 1.16 \\
 & 1024 & 10/10 & 1.00 ± 0.00 & 8.54 ± 2.23 \\
 & 2048 & 10/10 & 1.00 ± 0.00 & 12.20 ± 3.77 \\
 & 4096 & 10/10 & 1.00 ± 0.00 & 16.97 ± 4.35 \\
\midrule
\multirow{10}{*}{\parbox{1.3cm}{\gputamp{}}} & 8 & 5/10 & 14.40 ± 17.83 & 13.88 ± 7.29 \\
 & 16 & 10/10 & 2.90 ± 1.63 & 6.27 ± 2.57 \\
 & 32 & 10/10 & 4.60 ± 6.44 & 4.50 ± 2.31 \\
 & 64 & 10/10 & 1.20 ± 0.30 & \textbf{2.94 ± 0.56} \\
 & 128 & 10/10 & 1.10 ± 0.23 & 3.33 ± 0.26 \\
 & 256 & 10/10 & 1.00 ± 0.00 & 3.28 ± 0.47 \\
 & 512 & 10/10 & 1.00 ± 0.00 & 4.37 ± 0.55 \\
 & 1024 & 10/10 & 1.00 ± 0.00 & 6.88 ± 1.13 \\
 & 2048 & 10/10 & 1.00 ± 0.00 & 10.82 ± 2.41 \\
 & 4096 & 10/10 & 1.00 ± 0.00 & 21.50 ± 5.40 \\
\midrule
\multirow{10}{*}{\parbox{1.3cm}{\gputamp{} w. subgraph caching}} & 8 & 1/10 & 11.00 ± 0.00 & 29.05 ± 0.00 \\
 & 16 & 6/10 & 8.67 ± 12.12 & 10.04 ± 14.77 \\
 & 32 & 8/10 & 10.00 ± 10.37 & 7.96 ± 3.75 \\
 & 64 & 10/10 & 3.30 ± 1.22 & 8.72 ± 3.33 \\
 & 128 & 10/10 & 4.20 ± 2.82 & 11.14 ± 8.14 \\
 & 256 & 10/10 & 1.00 ± 0.00 & 1.46 ± 0.12 \\
 & 512 & 10/10 & 1.00 ± 0.00 & \textbf{\hl{1.45 ± 0.05}} \\
 & 1024 & 10/10 & 1.00 ± 0.00 & 1.75 ± 0.10 \\
 & 2048 & 10/10 & 1.00 ± 0.00 & 2.06 ± 0.04 \\
 & 4096 & 10/10 & 1.00 ± 0.00 & 3.17 ± 0.04 \\
\bottomrule
\end{tabular}

\end{small}
\captionof{table}{
\textbf{Full Results on Stick Button.} We present our results on Stick Button for different batch sizes. \textsc{Sampling} with a batch size of 8 happens to find a solution very quickly in 1/10 trials by pure chance.
}
\label{tab:stick-button-full-results}
\end{figure}

\lstdefinestyle{pythonstyle}{
language=Python, 
breaklines=true,
basicstyle=\ttfamily\footnotesize,
emphstyle=\bfseries\color{deepred}, 
emph={forward,forward_v},         
numbers=left,
numberstyle=\tiny\color{gray},
stepnumber=1,
numbersep=12pt,
tabsize=2,
stringstyle=\color{textgreen},
frame=none,                    
columns=fullflexible,
keepspaces=true,
xleftmargin=\parindent,
showstringspaces=false,
commentstyle=\color{deepgreen},
keywordstyle=\color{es-blue},
}

\begin{figure*}[ht]
\centering
\begin{lstlisting}[style=pythonstyle]
import torch
from jaxtyping import Float

from curobo.types.math import Pose


def curobo_pose_error(
    pose_a_mat4x4: Float[torch.Tensor, "b *h 4 4"], pose_b_mat4x4: Float[torch.Tensor, "b *h 4 4"]
) -> (Float[torch.Tensor, "b *h"], Float[torch.Tensor, "b *h"]):
    """
    Computes the translational and rotational errors between two poses using curobo.
    Used for computing end-effector pose errors for kinematic constraints.
    """
    # Flatten
    pose_a_flat = pose_a_mat4x4.view(-1, 4, 4)
    pose_b_flat = pose_b_mat4x4.view(-1, 4, 4)

    # Create curobo pose
    pose_a = Pose.from_matrix(pose_a_flat)
    pose_b = Pose.from_matrix(pose_b_flat)

    # Compute distance and unflatten
    p_dist_flat, quat_dist_flat = pose_a.distance(pose_b)
    p_dist = p_dist_flat.view(pose_a_mat4x4.shape[:-2])
    quat_dist = quat_dist_flat.view(pose_b_mat4x4.shape[:-2])
    return p_dist, quat_dist


def dist_from_bounds(
    vals: Float[torch.Tensor, "b *h d"],
    lower: Float[torch.Tensor, "d"],
    upper: Float[torch.Tensor, "d"],
) -> Float[torch.Tensor, "b *h"]:
    """
    Euclidean distance of values from the given lower and upper bounds. If within the bounds, returns 0.
    Used to check if objects are within the goal region, robot joint limits, etc.
    """
    diff_lower = lower - vals
    diff_upper = vals - upper
    diff_max = torch.maximum(diff_lower, diff_upper)
    diff_max = diff_max.clamp(min=0.0)
    dists = diff_max.norm(p=2, dim=-1)
    return dists


def obj_dist(goal_obj_position: Float[torch.Tensor, "b n 3"]) -> Float[torch.Tensor, "b"]:
    """
    Computes the total distance between all pairs of goal object positions.
    Can be used to minimize or maximize the distance between objects.
    """
    all_obj_dists = torch.cdist(goal_obj_position, goal_obj_position, p=2)  # (b, n, n)
    mask = torch.triu(torch.ones_like(all_obj_dists), diagonal=1) == 1
    obj_dists = all_obj_dists[mask].view(mask.shape[0], -1)  # reshape into num pairs
    dists_sum = obj_dists.sum(-1)
    return dists_sum
\end{lstlisting}
\captionof{lstlisting}{
\textbf{Example Cost Functions.} We present vectorized implementations for the following:
\texttt{curobo\_pose\_error}: computes the error between batches of poses, used in kinematic costs.
\texttt{dist\_from\_bounds}: calculates the distance from boundaries, used for verifying whether an object is within the goal region or whether configurations are within a robot's joint limits.
\texttt{obj\_dist}: computes the distance between object positions, employed in the soft cost to minimize object distance (see Figures~\ref{fig:ur5-min-obj-dist} and~\ref{fig:obstacle-blocks-soft}).
}
\label{appendix:example-cost-functions}
\end{figure*}

\end{document}